%% file: acl_latex.tex
\documentclass[11pt]{article}

\usepackage[preprint]{acl}

\usepackage{times}
\usepackage{latexsym}

\usepackage[T1]{fontenc}

\usepackage[utf8]{inputenc}

\usepackage{microtype}

\usepackage{inconsolata}

\usepackage{graphicx}

\usepackage{amsmath, amssymb}
\usepackage{bm}           
\usepackage{booktabs}     
\usepackage{multirow}
\usepackage{multicol}
\usepackage{array}
\usepackage{xcolor}
\usepackage{colortbl}
\usepackage{rotating}
\usepackage{geometry}
\usepackage{caption}
\usepackage{siunitx}
\usepackage{makecell}
\usepackage{placeins}
\usepackage{pdflscape}
\usepackage{pifont}

\usepackage{tikz}
\usepackage{xcolor}
\usepackage{float}   


\usetikzlibrary{arrows.meta, positioning, fit, backgrounds, shapes.geometric, calc, shadows.blur
}

\definecolor{Tcol}{RGB}{31,  97, 141}
\definecolor{Acol}{RGB}{30, 132,  73}
\definecolor{Vcol}{RGB}{175, 52,  35}
\definecolor{KDcol}{RGB}{118, 68, 138}
\definecolor{PUREbdr}{RGB}{30, 132, 73}
\definecolor{PUREfill}{RGB}{213,245,227}
\definecolor{ATTbdr}{RGB}{202,111, 30}
\definecolor{ATTfill}{RGB}{253,240,213}
\definecolor{GATEcol}{RGB}{44, 62, 80}
\definecolor{panelbg}{RGB}{247,248,252}
\definecolor{sepline}{RGB}{189,195,199}
\definecolor{regcol}{RGB}{85, 95,110}

\newcommand{\best}[1]{\textbf{#1}}
\newcommand{\pos}[1]{\textcolor{blue}{#1}}
\newcommand{\nega}[1]{\textcolor{red}{#1}}
\title{Mitigating Strong-Modality Collapse in Multimodal Learning via Inverted Asymmetric Fusion}

\author{
  \textbf{Mary Ogbuka Kenneth\textsuperscript{1}},
  \textbf{Foaad Khosmood\textsuperscript{2}},
  \textbf{Abbas Edalat\textsuperscript{1}}
\\
\\
  \textsuperscript{1} Algorithmic Human Development group, Department of Computing \\
  Imperial College London, United Kingdom\\
  \textsuperscript{2} Computer Engineering Department, California Polytechnic State University   \\
  San Luis Obispo, United States
\\
  \small{
    \textbf{Correspondence:} \href{mailto:m.kenneth22@imperial.ac.uk}{m.kenneth22@imperial.ac.uk}
  }
}

\begin{document}
\maketitle

\begin{abstract}
Fusing multiple modalities is expected to improve model performance. However, on the MultiHuSE dataset, early, late, and symmetric attention fusion often fail to outperform the best unimodal baseline (text). Pathway isolation of a symmetric attention fusion model reveals that the text-pathway accuracy drops from 74.9\% to 56.4\% after fusion in one such setting, indicating that the dominant modality can be degraded during integration. We term this strong-modality collapse and argue that it helps explain why some multimodal models fail to surpass unimodal baselines. We propose Inverted Asymmetric Fusion (IAF), which avoids forcing mutual attention across modalities. The dominant modality is preserved by passing through fusion unchanged, while weaker modalities attend to it as a contextual anchor. Before fusion, weaker modalities are strengthened using Modality-Aware Knowledge Distillation. We evaluate IAF on three benchmarks with different modality hierarchies: text-dominant datasets (MultiHuSE, UR-FUNNY) and an audio-visual-dominant dataset (MUStARD). Pathway isolation shows that IAF preserves the dominant modality’s internal accuracy at its unimodal ceiling across all tested configurations, whereas symmetric fusion degrades it by up to 18.5\% on MultiHuSE. IAF improves over the strongest unimodal baseline by up to 8.25\%.
\end{abstract}

\section{Introduction}
\label{sec:intro}
Early, late, and symmetric attention fusion form the standard toolkit for integrating text, audio, and video, resting on the intuition that combining heterogeneous signals should enable models to outperform any single modality alone \cite{Baltrusaitis2019MultimodalTaxonomy, Li2025MultimodalSurvey}. Yet on MultiHuSE \citep{Kenneth2025MultiHuSE:Emotions}, we observe that these approaches often fail to surpass a text-only baseline and can even degrade performance in some settings. This pattern is not isolated: prior work in affective computing and action recognition reports similar findings, where multimodal models trained with conventional fusion strategies underperform their strongest unimodal component despite having access to additional modalities \citep{Peng2022BalancedModulation, Wang2020WhatHard}.

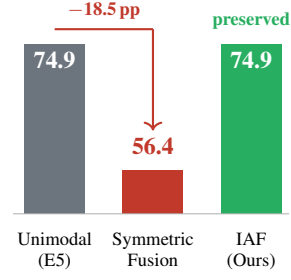
\begin{figure}[t]
\centering
\begin{tikzpicture}[font=\small]
  \definecolor{col1}{RGB}{108,117,125}
  \definecolor{col2}{RGB}{192,57,43}
  \definecolor{col3}{RGB}{39,174,96}


  \fill[col1] (0.10,0) rectangle (0.90,2.241);
  \fill[col2] (1.40,0) rectangle (2.20,0.576);
  \fill[col3] (2.70,0) rectangle (3.50,2.241);

  \draw[black!30](-0.05,0)--(3.60,0);

  \node[font=\small\bfseries,white]         at (0.50,2.041){74.9};
  \node[font=\small\bfseries,col2,above=2pt] at (1.80,0.576){56.4};
  \node[font=\small\bfseries,white]         at (3.10,2.041){74.9};

  \node[align=center,font=\scriptsize,below=3pt] at (0.50,0){Unimodal\\(E5)};
  \node[align=center,font=\scriptsize,below=3pt] at (1.80,0){Symmetric\\Fusion};
  \node[align=center,font=\scriptsize,below=3pt] at (3.10,0){IAF\\(Ours)};

  \draw[col2,thick](0.50,2.441)--(1.80,2.441);
  \draw[col2,thick,->](1.80,2.441)--(1.80,1.1);
  \node[col2,font=\scriptsize\bfseries,above] at (1.15,2.441){$-$18.5\,pp};

  \node[col3,font=\scriptsize\bfseries,above=2pt] at (3.10,2.241){preserved};

\end{tikzpicture}
\caption{Text-pathway accuracy on MultiHuSE (E5 encoder). Symmetric fusion degrades the dominant modality by 18.5\,pp; IAF preserves it
(Table~\ref{tab:pathway_summary}).}
\label{fig:collapse_motivation}
\end{figure}

Prior work attributes such failures to optimisation dynamics: modalities overfit and generalise at different rates, causing training to gravitate toward whichever modality converges fastest \cite{Chaudhuri2025ACollapse, Huang2022ModalityProvably, Wang2020WhatHard}. We extend this view by identifying an additional architectural mechanism that may contribute to the problem. Pathway isolation analysis of a symmetric attention model on MultiHuSE shows that the dominant text pathway drops from 74.9\% to 56.4\% after fusion (Figure \ref{fig:collapse_motivation}). 

This decline suggests that symmetric cross-attention forces the dominant modality to attend to weaker, noisier signals, degrading the representation that initially carried the most predictive information. We term this phenomenon \textit{strong-modality collapse:} a failure mode in which fusion undermines the model’s strongest modality.

This diagnosis suggests an architectural alternative. If symmetric attention contributes to this degradation, the solution lies in restructuring the fusion pathway rather than solely reweighting gradients or losses~\cite{Wei2025On-the-FlyLearning, Fan2023PMR:Learning, Javaloy2022MitigatingOptimization}, which adjust training dynamics while leaving the cross-attention topology intact. To this end, we propose Inverted Asymmetric Fusion (IAF), which imposes an explicit structural hierarchy. In IAF, the dominant modality bypasses cross-modal attention entirely to preserve its representation, while weaker modalities attend to it as a contextual anchor rather than as standalone classifiers. Additionally, a modality-aware knowledge distillation stage \cite{Hinton2015DistillingNetwork} precedes fusion, strengthening weaker modality representations so they can contribute a more informative complementary signal to the dominant pathway.

Testing across inverted modality hierarchies is essential to establish that IAF's structural approach generalises beyond any single dominance configuration. We therefore evaluate on MultiHuSE \citep{Kenneth2025MultiHuSE:Emotions} and UR-FUNNY \citep{KamrulHasan2019UR-FUNNY:Humor}, where text is dominant, and MUStARD \citep{Castro2019TowardsPaper}, where acoustic and visual cues carry more predictive signal. Our contributions are as follows:

\begin{itemize}
    \item We identify strong-modality collapse as an architectural failure mode, distinct from optimisation-based methods that reweight gradients while preserving symmetric cross-attention, and provide pathway-level evidence in symmetric fusion.

    \item We propose IAF, a fusion framework that prevents strong-modality collapse by design and preserves the dominant modality’s pathway accuracy at its unimodal ceiling across all tested configurations.
  
    \item We introduce a Modality-Aware Knowledge Distillation pipeline that adapts distillation to each dataset’s modality hierarchy, strengthening weaker encoders prior to fusion.
    
    \item We demonstrate cross-hierarchy robustness across 3 benchmarks with differing dominance structures, improving on prior baselines on UR-FUNNY and MUStARD.
\end{itemize}

\section{Related Work}
\label{sec:related_work}

\subsection{Multimodal Fusion Architectures}
Multimodal fusion has evolved from simple concatenation and prediction averaging to learned attention mechanisms \cite{Baltrusaitis2019MultimodalTaxonomy, PuLiang2021MULTIBENCH:Learning}. Tensor-based approaches explicitly model cross-modal interactions via outer products \cite{Zadeh2017TensorAnalysis}, with later work reducing computational cost through low-rank factorisation \cite{Liu2018EfficientFactors}. The advent of attention-based fusion, led by the Multimodal Transformer (MulT) \cite{Tsai2020MultimodalSequences} established symmetric cross-attention as the dominant paradigm. Subsequent improvements incorporated pretrained language models \cite{Rahman2020IntegratingTransformers}, bottleneck token constraints \cite{Nagrani2021AttentionFusion}, and modality subspace disentanglement \cite{Hazarika2020MISA:Analysis}.

While these designs capture rich cross-modal dynamics, they apply attention uniformly, regardless of the relative modality strengths \cite{Feng2024Knowledge-GuidedAnalysis}. Forcing a dominant modality to symmetrically attend to weaker, noisier signals can corrupt its internal representation. Recent work has begun to address this by using text to guide multimodal fusion in affective tasks~\cite{Zhang2025Text-guidedUnderstanding}, though such designs typically retain mutual attention rather than fully shielding the dominant stream from cross-modal interference. This vulnerability motivates our inverted asymmetric approach, which preserves the dominant modality while using weaker streams as contextual anchors.

\subsection{Modality Collapse and Pathway Analysis}
\label{sec:collapse}
Representational degradation in fusion is one instance of a broader failure mode: modality collapse. Empirical studies show multimodal models often exploit only a subset of their input modalities \cite{Gong2025MultimodalLearning, Kenneth2024SystematicClassification, Zhang2024MultimodalAdaptation}. Because different modalities overfit and generalise at uneven rates, training tends to favour the fastest-converging modality \cite{Wang2020WhatHard, Wu2022CharacterizingNetworks}. Theoretical analyses attribute this behaviour to conflicting gradient directions during joint optimisation \cite{Huang2022ModalityProvably, Chaudhuri2025ACollapse}. 

To mitigate modality imbalance, prior work has proposed optimisation-based strategies such as on-the-fly gradient modulation (OGM-GE)~\cite{Wei2025On-the-FlyLearning}, Pareto-based integration~\cite{Wei2024MMPareto:Assistance}, prototype-guided clustering (PMR)~\cite{Fan2023PMR:Learning}, and impartial multitask optimisation~\cite{Javaloy2022MitigatingOptimization}. These methods address imbalance by adjusting gradients or training objectives, but generally retain the same fusion topology, leaving dominant representations vulnerable to cross-modal interference. In contrast, our work takes a complementary architectural approach. IAF restructures the fusion pathway to limit such interference by design. However, structural shielding alone is insufficient unless weaker encoders are first strengthened, motivating the pre-fusion distillation stage.

\subsection{Knowledge Distillation in Multimodal Learning}
\label{sec:kd}
Knowledge distillation (KD) transfers generalisation by aligning a student's predictions with a high-capacity teacher \cite{Hinton2015DistillingNetwork, Gou2021KnowledgeSurvey}. Cross-modal distillation extends this across domain boundaries and has proven effective for depth, optical flow, action recognition, and emotion analysis \cite{Gupta2016CrossTransfer, Thoker2019Cross-ModalRecognition, Albanie2018EmotionWild}. In multimodal fusion, KD is commonly used within the fusion stage for compression, dropout robustness, missing-modality learning, or representational alignment \cite{Fang2021CompressingDistillation, Wang2023EfficientVLM:Pruning, Wang2020MultimodalDistillation, Wei2023MMANet:Learning, Li2023DecoupledRecognition, Lin2024Multi-TaskAnalysis}. By contrast, our Modality-Aware Knowledge Distillation (MAKD) is a pre-fusion stage that uses only the empirically dominant unimodal modality as teacher and is designed specifically to strengthen weaker encoders before asymmetric integration. A detailed design comparison is provided in Appendix~\ref{app:kd_comparison}.

\subsection{Humor and Sarcasm Detection}
\label{sec:tasks}
Humor and sarcasm benchmarks provide a useful testbed for modality imbalance because their predictive hierarchies differ substantially across datasets. UR-FUNNY \cite{KamrulHasan2019UR-FUNNY:Humor} and MultiHuSE \cite{Kenneth2025MultiHuSE:Emotions, Kenneth2024ARecognition, Kenneth2025ExplainingAnalysis} are text-dominant, whereas MUStARD \cite{Castro2019TowardsPaper} places greater weight on acoustic and visual cues. This contrast makes the domain well suited for evaluating whether a fusion architecture can preserve strong modalities without sacrificing cross-modal contribution.


\section{Methodology}
\label{sec:method}

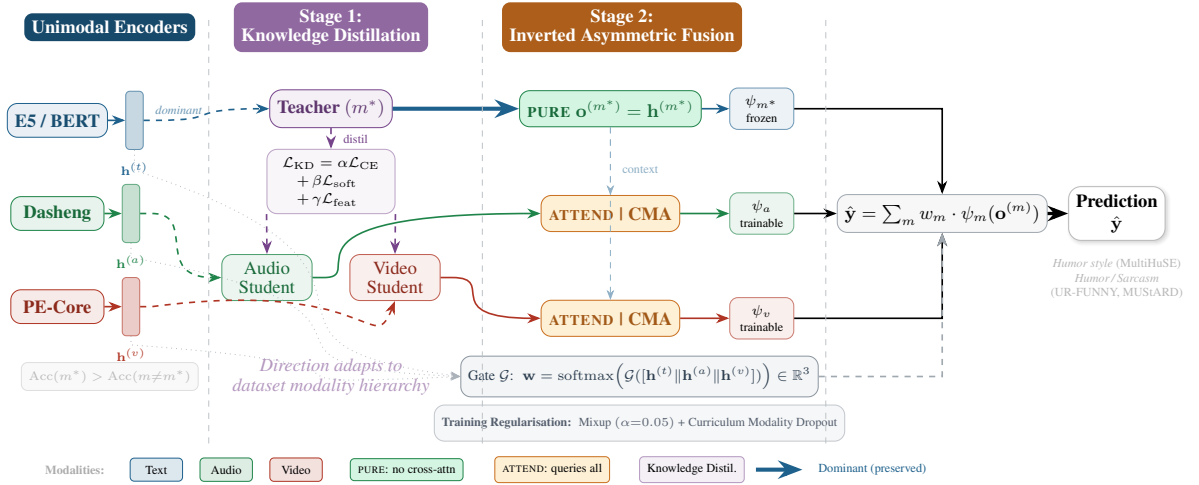
\begin{figure*}[t]
  \centering
  \resizebox{\textwidth}{!}{%
    \input{methodology_figure.tex}%
  }
  \caption{Overview of the proposed two-stage framework. Stage~1 performs Modality-Aware Knowledge Distillation, and Stage~2 applies the Inverted Asymmetric Fusion.}
  \label{fig:methodology}
\end{figure*}

Our framework comprises two sequential stages (Figure~\ref{fig:methodology}). \textbf{Stage 1} applies \textit {Modality-Aware Knowledge Distillation} to strengthen weaker modalities prior to fusion, adapting the strategy to each dataset's modality hierarchy. \textbf{Stage 2} performs \textit{Inverted Asymmetric Fusion} (IAF), integrating streams within a strict hierarchy that shields the dominant modality from cross-modal interference and prevents strong-modality collapse.

\subsection{Problem Formulation}
\label{ssec:formulation}

Let $\mathbf{f}^{(t)} \in \mathbb{R}^{d_t}$,
$\mathbf{f}^{(a)} \in \mathbb{R}^{d_a}$, and
$\mathbf{f}^{(v)} \in \mathbb{R}^{d_v}$ denote pre-extracted feature vectors for text, audio, and video modalities of a single utterance, respectively. Each is projected to a shared $d{=}512$ space via a modality-specific encoder $\phi_m$, yielding
$\mathbf{h}^{(m)} = \phi_m(\mathbf{f}^{(m)}) \in \mathbb{R}^{512}$.
The \textbf{dominant modality} $m^{*}$ is defined as $m^{*} = \arg\max_{m} \;\text{Accuracy}(\phi_m)$ on the validation set.

\subsection{Stage 1: Modality-Aware Knowledge Distillation}
\label{ssec:distillation}

The dominant modality $m^{*}$ acts as a teacher to elevate weaker modalities $m \neq m^{*}$ before fusion. 

\subsubsection{Distillation Loss (\texorpdfstring{$\mathcal{L}_{\text{KD}}$}{LKD}).}
Each student $\mathcal{S}$ is trained against a frozen teacher $\mathcal{T}$ (logits $\mathbf{z}^T \in \mathbb{R}^C$, bottleneck features $\mathbf{g}^T \in \mathbb{R}^{512}$) using:
\begin{equation}
\mathcal{L}_{\text{KD}} \;=\;
  \alpha\,\mathcal{L}_{\text{CE}}
  + \beta\,\mathcal{L}_{\text{soft}}
  + \gamma\,\mathcal{L}_{\text{feat}}
\label{eq:kd_loss}
\end{equation}
The loss terms are: hard‑label cross‑entropy \(\mathcal{L}_{\text{CE}}\), temperature‑scaled KL divergence \(\mathcal{L}_{\text{soft}}\) (with \(T = 3.5\)), and bottleneck feature alignment \(\mathcal{L}_{\text{feat}}\):
\begin{align}
\mathcal{L}_{\text{CE}} 
&= -\sum_{c=1}^{C} \mathbf{1}[y = c]\, 
    \log\!\bigl(\text{softmax}(\mathbf{z}^{S})_c\bigr) \\[4pt]
\mathcal{L}_{\text{soft}} 
&= T^{2}\,\text{KL}\!\left(
        \sigma\!\left(\frac{\mathbf{z}^{S}}{T}\right)
        \,\middle\|\,
        \sigma\!\left(\frac{\mathbf{z}^{T}}{T}\right)
    \right) \\[4pt]
\mathcal{L}_{\text{feat}} 
&= \left\|\mathbf{g}^{S} - \mathbf{g}^{T}\right\|_2^{2}
\end{align}
where $\sigma$ denotes the softmax function. We use fixed coefficients $\alpha=0.4$, $\beta=0.35$, and $\gamma=0.25$, selected by manual tuning on MultiHuSE and held constant across all datasets and encoder combinations. A sensitivity sweep of coefficients over four configurations shows that fusion accuracy varies by at most 1.20 percentage points, with both the default and equal-weight settings achieving the best result; see Appendix~\ref{app:sensitivity} for details.

\paragraph{Weighted Dual-Teacher Distillation (MUStARD).}
When both audio and video dominate (MUStARD), students are supervised by weighted dual teachers. The student is jointly supervised by a primary teacher $\mathcal{T}_1$ and a secondary teacher $\mathcal{T}_2$, whose losses are combined via a scalar weight $w_1$:
\begin{equation}
\mathcal{L}_{\text{dual}} = w_1\,\mathcal{L}_{\text{KD}}(\mathcal{T}_1)
  + (1 - w_1)\,\mathcal{L}_{\text{KD}}(\mathcal{T}_2)
\label{eq:dual_loss}
\end{equation}
where $w_1$ is tuned per configuration ($w_1=0.8$ for audio-primary, $w_1=0.2$ for video-primary).

\paragraph{Cross-Architecture Distillation (UR-FUNNY).}
On UR-FUNNY, the narrow unimodal performance gap limits the representational surplus available for cross-modal transfer. This motivated a same-modality, cross-architecture distillation: high-capacity static teachers (E5, Dasheng, PE-Core) distill into BiLSTM students on benchmark features (GloVe, COVAREP, OpenFace), isolating IAF's architectural contribution. The same three-term loss in Eq.~\eqref{eq:kd_loss} applies, with teacher features $\mathbf{g}^{T}$ projected to 512-dim prior to alignment.

With all modality encoders operating at an enhanced representation, Stage~2 integrates these streams using IAF.

\subsection{Stage 2: Inverted Asymmetric Fusion}
\label{ssec:fusion}

\subsubsection{CrossModal Anchor (CMA)}
The \emph{CrossModal Anchor} module serves as the core attention primitive in our architecture. Given a query-modality vector $\mathbf{h} \in \mathbb{R}^{512}$ and a context set $\mathbf{C} \in \mathbb{R}^{K \times 512}$ containing $K$ other modalities, the module computes:
\begin{align}
\hat{\mathbf{h}} &= \text{MHA}(\mathbf{h},\, \mathbf{C},\, \mathbf{C})
  \label{eq:mha} \\
\mathbf{h}' &= \text{LayerNorm}\!\left(
  \mathbf{h} + \text{Dropout}(\hat{\mathbf{h}})
\right)
  \label{eq:residual}
\end{align}
where $\text{MHA}(\cdot)$ denotes multi-head attention \cite{Vaswani2017AttentionNeed} with $d_\text{model}{=}512$ and $H{=}8$ heads. The residual connection Eq.~\eqref{eq:residual} preserves the query modality's original representation.

\subsubsection{Structural Hierarchy: \textsc{Pure} vs.\ \textsc{Attend}}
Let $m^{*}$ denote the dominant modality (or modalities). IAF enforces the following constraint on the forward pass:
{\footnotesize
\begin{equation}
\mathbf{o}^{(m)} =
\begin{cases}
  \mathbf{h}^{(m)} & \text{if } m = m^{*} \; (\textsc{pure}) \\[4pt]
  \text{CMA}\!\left(
    \mathbf{h}^{(m)},\;
    [\mathbf{h}^{(j)}]_{j \neq m}
  \right) & \text{if } m \neq m^{*} \; (\textsc{attend})
\end{cases}
\label{eq:hierarchy}
\end{equation}
}

The \textsc{pure} pathway remains unmodified. \textsc{Attend} pathways query all other modalities (including the shielded dominant) as contextual filters. On MultiHuSE/UR-FUNNY, text is \textsc{pure}; on MUStARD, audio/video.

\subsubsection{Adaptive Gating and Final Classification}
Each modality produces pathway logits via a dedicated classifier head $\psi_m$. For the \textsc{pure} modality, $\psi_{m^*}$ is a frozen copy of its standalone classifier, preserving its decision boundary. For \textsc{attend} modalities, $\psi_m$ is a trainable linear head suited to their contextually-enriched representations.

A gating network $\mathcal{G}$ computes sample-adaptive weights from pre-attention features:
\begin{equation}
\mathbf{w} = \text{softmax}\!\left(
  \mathcal{G}\!\left([\mathbf{h}^{(t)} \,\|\, \mathbf{h}^{(a)} \,\|\, \mathbf{h}^{(v)}]\right)
\right)
\label{eq:gate}
\end{equation}
where $\mathbf{w} \in \mathbb{R}^3$ denotes the modality‑gate weights, and $\mathcal{G}$ is a two‑layer MLP with batch normalisation and dropout. The final prediction is the weighted sum of per-pathway logits:
\begin{equation}
\hat{\mathbf{y}} = \sum_{m \in \{t,a,v\}} w_m \cdot \psi_m(\mathbf{o}^{(m)})
\label{eq:final_logits}
\end{equation}
Gating on pre-attention features preserves the textsc{pure}/\textsc{attend} separation.

\subsection{Training Regularisation}
\label{ssec:regularisation}
We apply two complementary regularisation strategies during fusion training.

\subsubsection{Mixup}
For each mini-batch, we sample a mixing coefficient $\lambda \sim \text{Beta}(\alpha_{\text{mix}},\, \alpha_{\text{mix}})$ with $\alpha_{\text{mix}}{=}0.05$ and interpolate pairs of feature 
vectors and their labels:
{\footnotesize
\begin{equation}
\tilde{\mathbf{h}}_i^{(m)} = \lambda\,\mathbf{h}_i^{(m)} +
  (1-\lambda)\,\mathbf{h}_{\sigma(i)}^{(m)}, \quad
\tilde{y}_i = (\lambda,\, y_i,\, y_{\sigma(i)})
\end{equation}
}
where $\sigma$ is a random permutation of batch indices. Mixup discourages overconfident predictions, which is critical for preventing overfitting (ablation in Appendix~\ref{app:reg_ablation})

\subsubsection{Curriculum Modality Dropout}
To prevent over-reliance on any single modality, we apply a two-phase stochastic masking schedule: all modalities are present during warm-up (epochs~$<$15), after which each sample is assigned one of seven masking configurations (trimodal $p=0.40$, each bimodal pair $p=0.10$, each unimodal $p=0.10$), with masked modalities zeroed before the forward pass.

\subsection{Pathway Analysis}
\label{ssec:pathway}
To verify dominant pathway preservation, we isolate each modality's internal contribution by routing $\mathbf{h}^{(m)}$ directly through $\psi_m$, bypassing gating.  For the \textsc{pure} modality $m^{*}$:
\begin{equation}
\text{Accuracy}_{\text{path}}^{(m^*)} = \text{Accuracy}_{\text{uni}}^{(m^*)}
\label{eq:doNoharm}
\end{equation}

That is, $m^{*}$ pathway accuracy equals its standalone unimodal performance. Any deviation from this equality indicates cross-modal interference. Symmetric fusion violates this equality; IAF enforces it by construction. For \textsc{attend} modalities, pathway accuracy is expected to drop below their unimodal baselines, as their representations are no longer optimised for standalone classification but instead tuned to provide contextual signal.


\section{Experiments}
\label{sec:experiments}
We describe the datasets, feature extraction procedures, baseline models, and implementation details.

\subsection{Datasets}
\label{ssec:datasets}
We evaluate on three multimodal benchmarks for humor and sarcasm detection, chosen to span contrasting modality hierarchies and task complexity (binary and multiclass classification).

\paragraph{MultiHuSE}~\cite{Kenneth2025MultiHuSE:Emotions} is a 5-class humor style classification dataset comprising 2,407 video utterances from 50 diverse actors, including 1,463 unique text scripts and 943 re-performances. Standard random splits risk data leakage across performances of the same script, so we introduce a strict 5-fold cross-validation protocol grouped by unique text ID, ensuring all performances of a given script appear exclusively in either training or evaluation. All baseline models are re-implemented and re-evaluated under this protocol. Modality hierarchy: Text $\gg$ Audio $\gg$ Video.

\paragraph{UR-FUNNY}~\cite{KamrulHasan2019UR-FUNNY:Humor} is a binary humor detection dataset of 16,514 context--punchline pairs drawn from TED talks, using the standard train/dev/test split ($N_{\text{test}}{=}994$). Following the original benchmark, we use sequential features GloVe (text), COVAREP (audio), and OpenFace (video) throughout, isolating our architectural contribution from feature quality and enabling direct comparison with prior work using the same feature set. Modality hierarchy: Text $>$ Audio $\approx$ Video.

\paragraph{MUStARD}~\cite{Castro2019TowardsPaper} is a binary sarcasm detection dataset of 690 balanced utterances drawn from TV sitcoms, using 5-fold cross-validation ($N_{\text{test}}{=}138$ per fold) from the original paper. Modality hierarchy: Audio $\approx$ Video $>$ Text, is the clearest inversion relative to MultiHuSE and motivates our cross-hierarchy evaluation.

\subsection{Feature Extraction}
\label{ssec:features}
All features are extracted offline using frozen pre-trained encoders and serialised to disk prior to training. This design decouples representation quality from the fusion architecture, ensuring that observed performance differences reflect architectural choices rather than effects of encoder fine-tuning.

\paragraph{MultiHuSE and MUStARD.}
For text, we evaluate two encoders: BERT-based-uncased~\cite{Devlin2019BERT:Understanding} (768-d) and Multilingual-E5-large-instruct~\cite{Wang2024MultilingualReport} (1024-d). Using both allows us to verify that our findings are not specific to a single text backbone. For audio, we use Dasheng-0.6B~\cite{Dinkel2024ScalingClassification} (1280-d), a large-scale general-purpose audio encoder. For video, we evaluate two encoders: VideoMAE~\cite{Tong2022VideoMAE:Pre-Training} (768-d) and PE-Core~\cite{Bolya2025PerceptionNetwork} (1024-d), providing coverage of both masked autoencoder and contrastive video representations.

\paragraph{UR-FUNNY.}
We use GloVe~\cite{Pennington2014GloVe:Representation} (text), 
COVAREP~\cite{Degottex2014COVAREP:Technologies} (audio), and 
OpenFace~\cite{Baltrusaitis2018OpenFaceToolkit} (video), consistent with the original benchmark. The role of modern encoders as distillation teachers within this established feature setting is described in Section~\ref{ssec:distillation}.

\subsection{Implementation Details}
\label{ssec:impl}

\paragraph{Infrastructure.}
Feature extraction is performed on an HPC cluster (NVIDIA A16, 15GB VRAM), while all subsequent training is conducted on a consumer laptop (NVIDIA GeForce RTX~3050, 4GB VRAM). All primary results use a fixed random seed of 999. For MultiHuSE and MUStARD, results are reported as mean $\pm$ SD over 5-fold cross-validation; for UR-FUNNY, we additionally report a five-seed analysis using seeds 42, 100, 123, 999, and 600. The unimodal pre-evaluation used to identify the dominant modality adds minimal overhead: on our hardware, training a single unimodal encoder takes less than 15 minutes, compared with a few hours for full fusion training.

\paragraph{Model Parameters.} Pre-trained encoders used for feature extraction have the following parameter counts: BERT-base (110M), E5-multilingual (560M), Dasheng (600M), VideoMAE-base (94M), and PE-Core (320M visual encoder + 310M language tower). These encoders are frozen throughout; no encoder parameters are updated during distillation or fusion training.

\paragraph{Distillation and Fusion.} We used AdamW for distillation, CosineAnnealingLR for fusion, a batch size of 64, Mixup with $\alpha_{\text{mix}} = 0.05$, a curriculum dropout warm-up of 15 epochs, and a CrossModal Anchor with $d_{\text{model}} = 512$ and $H = 8$ heads. Full hyperparameters are provided in Appendix~\ref{app:hyperparams}.

\paragraph{Evaluation.} We report accuracy and macro-averaged F1. For MultiHuSE and MUStARD, results are reported as mean $\pm$ SD over 5 folds. For UR-FUNNY, we use the standard train/dev/test split; the result at seed~999 is reported in Table~\ref{tab:urfunny_main} for direct comparability with prior published baselines, and a multi-seed analysis is reported in Section~\ref{ssec:results}.

\subsection{Baselines}
\label{ssec:baselines}
We compare against three internal baselines trained on identical features under identical hyperparameters: (1)~\textbf{Early Fusion}: feature concatenation followed by an MLP; (2)~\textbf{Late Fusion}: weighted averaging of unimodal prediction vectors; (3)~\textbf{Symmetric Fusion}: mutual cross-attention across all three modalities using the same distilled features as IAF, isolating the inverted asymmetric design. Comparisons with prior published work appear in the results tables in Section~\ref{ssec:results}.

\section{Results and Analysis}
\label{ssec:results}
Table~\ref{tab:unimodal_all} reports unimodal performance. Distillation consistently improves student models (gains of 0.88–6.23\%, mean 2.44\%), with the largest gain on UR-FUNNY's COVAREP audio ($+$6.23\%), highlighting the benefit of cross-architecture distillation. These distilled students, together with their teachers, serve as the inputs to IAF and symmetry fusion.

\begin{table}[ht]
\centering

{\setlength{\tabcolsep}{4pt}
\resizebox{\columnwidth}{!}{
\begin{tabular}{lcccc}
\toprule
\textbf{Modality / Feature} & \textbf{Base Acc(\%)} & \textbf{Base F1(\%)} & \textbf{Dist. Acc(\%)} & \textbf{Dist. F1(\%)}\\
\midrule
\multicolumn{5}{l}{\textit{\textbf{MultiHuSE} — 5-class humor style | Text $\gg$ Audio $\gg$ Video}} \\
\midrule
Text: BERT           & 68.26$\pm$2.25 & 68 & 70.21 $\pm$ 2.82     & 70 \\
Text: E5$^\star$ & \best{74.86$\pm$3.08} & \best{75}  & --  & --  \\
Audio: Dasheng       & 60.45$\pm$1.58 & 60  & 62.90 $\pm$ 1.83    &  63 \\
Video: PE-Core       & 43.66$\pm$1.29 & 43  & 45.57 $\pm$ 1.22    & 46 \\
Video: VideoMAE      & 40.13$\pm$0.73 & 40  & 41.01 $\pm$ 1.60    & 41 \\
\midrule
\multicolumn{5}{l}{\textit{\textbf{UR-FUNNY} — binary humor | Text $>$ Audio $\approx$ Video}} \\
\midrule
Text: GloVe$^\star$     & 62.47  & 64  & \best{65.19} & \best{65} \\
Audio: COVAREP          & 57.55  & 48  & 63.78  & 64 \\
Video: OpenFace         & 57.75  & 58  & 58.95  & 59 \\
\midrule
\multicolumn{5}{l}{\textit{\textbf{MUStARD} — binary sarcasm | Audio $\approx$ Video $>$ Text}} \\
\midrule
Text: BERT          & 69.42$\pm$2.80 & 69  & 72.03$\pm$ 2.49  & 72 \\
Text: E5            & 70.29$\pm$2.00 & 70  & 72.32$\pm$ 2.26  & 72 \\
Audio: Dasheng$^\star$      & \best{78.12$\pm$5.45} & \best{78}  & --  & --  \\
Video: PE-Core      & 77.10$\pm$3.13 & 77  & --  & -- \\
Video: VideoMAE     & 74.49$\pm$1.16 & 74  & --  & -- \\
\bottomrule
\end{tabular}
}}
\caption{Unimodal results across all three benchmarks. \textit{Base}: raw pre-trained; \textit{Dist.}: knowledge-distilled. $^\star$ = dominant modality. MultiHuSE and MUStARD: mean\,$\pm$\,SD over 5-fold CV; UR-FUNNY: single split.}

\label{tab:unimodal_all}
\end{table}

Table~\ref{tab:multihuse_main} presents trimodal fusion results on MultiHuSE. Early and late fusion fail to reliably surpass the dominant unimodal text baseline, with early fusion degrading sharply in the E5+VideoMAE setting ($-10.55\%$), suggesting that naive concatenation may disrupt strong text representations when combined with noisier video inputs. Symmetric fusion also fails to recover the E5 baseline in both E5 configurations ($-$2.49\%, $-$0.54\%), indicating that expressive attention alone is insufficient to preserve the dominant modality. IAF is the only architecture that consistently exceeds the unimodal ceiling across all four configurations, attaining gains of 2.06–6.48\%. Paired Wilcoxon signed-rank tests across the five folds show that IAF significantly outperforms symmetric fusion in the PE-Core configurations (BERT: $W{=}0$, $p{=}0.031$; E5: $W{=}0$, $p{=}0.031$; one-tailed), a result further supported by paired $t$-tests ($p{=}0.023$ and $p{=}0.005$, respectively); Here, $W{=}0$ indicates IAF outperformed symmetric fusion on every fold.

\begin{table}[ht]
\centering
{\setlength{\tabcolsep}{4pt}
\resizebox{\columnwidth}{!}{
\begin{tabular}{lllccc}
\toprule
\textbf{Text} & \textbf{Video} & \textbf{Architecture}
  & \textbf{Acc} & \textbf{F1} & \textbf{$\Delta$} \\
\midrule
\multicolumn{6}{c}{\textit{Audio: Dasheng used across all configurations}} \\
\midrule
\multicolumn{6}{l}{\textit{Text: BERT (base 68.26\%)}} \\
\multirow{4}{*}{BERT} & \multirow{4}{*}{VideoMAE}
  & Early Fusion            & 67.22 {\small$\pm$2.00} & 67 & \nega{$-1.04$} \\
 && Late Fusion             & 68.05 {\small$\pm$2.12} & 68 & \nega{$-0.21$} \\
 && Symmetric Fusion        & 69.42 {\small$\pm$1.48} & 69 & \pos{$+1.16$} \\
 && \textbf{Proposed (IAF)} & \best{73.16 {\small$\pm$2.59}} & \best{72} & \pos{$+4.90$} \\
\cmidrule{1-6}
\multirow{4}{*}{BERT} & \multirow{4}{*}{PE-Core}
  & Early Fusion            & 70.38 {\small$\pm$2.57} & 70 & \pos{$+2.12$} \\
 && Late Fusion             & 68.63 {\small$\pm$2.36} & 68 & \pos{$+0.37$} \\
 && Symmetric Fusion        & 72.00 {\small$\pm$1.20} & 70 & \pos{$+3.74$} \\
 && \textbf{Proposed (IAF)} & \best{74.74 {\small$\pm$1.71}} & \best{74} & \pos{$+6.48$} \\
\midrule
\multicolumn{6}{l}{\textit{Text: E5 (base 74.86\%)}} \\
\multirow{4}{*}{E5} & \multirow{4}{*}{VideoMAE}
  & Early Fusion            & 64.31 {\small$\pm$1.51} & 64 & \nega{$-10.55$} \\
 && Late Fusion             & 71.25 {\small$\pm$4.07} & 71 & \nega{$-3.61$} \\
 && Symmetric Fusion        & 72.37 {\small$\pm$2.73} & 72 & \nega{$-2.49$} \\
 && \textbf{Proposed (IAF)} & \best{76.92 {\small$\pm$3.53}} & \best{76} & \pos{$+2.06$} \\
\cmidrule{1-6}
\multirow{4}{*}{E5} & \multirow{4}{*}{PE-Core}
  & Early Fusion            & 76.28 {\small$\pm$2.91} & 76 & \pos{$+1.42$} \\
 && Late Fusion             & 71.54 {\small$\pm$4.20} & 71 & \nega{$-3.32$} \\
 && Symmetric Fusion        & 74.32 {\small$\pm$3.03} & 74 & \nega{$-0.54$} \\
 && \textbf{Proposed (IAF)} & \best{78.06 {\small$\pm$3.82}} & \best{78} & \pos{$+3.20$} \\
\bottomrule
\end{tabular}
}}
\caption{Trimodal fusion results on MultiHuSE. $\Delta$ = gain over the strongest unimodal text baseline (BERT\,=\,68.26\%; E5\,=\,74.86\%). All metrics in \%.}
\label{tab:multihuse_main}
\end{table}

Table~\ref{tab:urfunny_main} compares IAF against prior work and our internal baselines on UR-FUNNY. IAF achieves 70.72\% accuracy, surpassing the strongest prior method (MISA, 68.60\%) by 2.12\% and symmetric fusion by 1.81\%. Interestingly, symmetric fusion (68.91\%) outperforms previously published methods, demonstrating that strengthening weaker modalities before fusion is beneficial even without architectural asymmetry. To assess initialisation sensitivity, we evaluated IAF over five random seeds (42, 100, 123, 600, 999), obtaining a mean accuracy of $69.70 \pm 0.61\%$ (min $69.32\%$; seed~999 is reported in Table~\ref{tab:urfunny_main} for comparability with prior work). A one-sample $t$-test shows that this mean significantly exceeds both symmetric fusion ($t(4){=}2.90$, $p{=}0.022$, one-tailed) and MISA ($t(4){=}4.03$, $p{=}0.008$, one-tailed), and each seed individually outperforms both baselines.

\begin{table}[ht]
\centering

{\setlength{\tabcolsep}{3pt}
\resizebox{\columnwidth}{!}{
\begin{tabular}{lcccc}
\toprule
\textbf{Method} & \textbf{Context} & \textbf{Acc (\%)} & \textbf{F1 (\%)} & \textbf{$\Delta_{\text{uni}}$} \\
\midrule
\multicolumn{5}{l}{\textit{Prior published methods}} \\
C-MFN~\cite{KamrulHasan2019UR-FUNNY:Humor}           & \texttimes & 64.47 & -- & -- \\
C-MFN~\cite{KamrulHasan2019UR-FUNNY:Humor}           & \checkmark & 65.23 & -- & -- \\
AGM(Late fusion)~\cite{Li2023BoostingModulation}     & \texttimes & 65.97 & -- & -- \\
AGM(Early fusion)~\cite{Li2023BoostingModulation}    & \texttimes & 66.07 & -- & -- \\
MULTIBENCH~\cite{PuLiang2021MULTIBENCH:Learning}     & \checkmark & 66.70 & -- & -- \\
BCFNet~\cite{Deng2025BCFNet:Detection}               & \texttimes & 66.78 & 68 & -- \\
HF~\cite{Choube2020PunchlineFusion}                  & \checkmark & 67.84 & 69 & -- \\
MISA (GloVe)~\cite{Hazarika2020MISA:Analysis}        & \texttimes & 68.60 & -- & -- \\
\midrule
\multicolumn{5}{l}{\textit{Our internal baselines --- Text: GloVe (base 62.47\%)}} \\
Early Fusion                     & \checkmark & 65.19 & 66 & \pos{$+2.72$} \\
Late Fusion                      & \checkmark & 64.19 & 65 & \pos{$+1.72$} \\
Symmetric Fusion                 & \checkmark & 68.91 & 69 & \pos{$+6.44$} \\
\textbf{Proposed (IAF)}         & \checkmark & \best{70.72}$^{\dagger}$  & 71 & \pos{$+8.25$} \\
\bottomrule
\end{tabular}
}}
\caption{Results on UR-FUNNY. All methods use
GloVe/COVAREP/OpenFace features~\cite{KamrulHasan2019UR-FUNNY:Humor}. $\Delta_{\text{uni}}$ = gain over the GloVe baseline (62.47\%) for our internal baselines. Context: \checkmark\,= punchline + context; \texttimes\,= punchline only. $^{\dagger}$Seed~999; mean $69.70\pm0.61\%$ over 5 seeds.}
\label{tab:urfunny_main}
\end{table}

Table~\ref{tab:mustard_main} reports results on MUStARD. Symmetric fusion already improves substantially over the strongest unimodal audio baseline (+3.47\% and +4.34\% for BERT and E5, respectively), and IAF yields further gains in both text-encoder settings. In particular, IAF (E5) attains 83.33\% accuracy, exceeding the state-of-the-art method (MHA with RoBERTa, 79.32\%) by 4.01\%. We note that the IAF-over-symmetric margin on MUStARD (0.87--1.02 pp) falls within one standard deviation of both models and should be interpreted accordingly. Paired Wilcoxon signed-rank tests are consistent with this pattern:  neither BERT+PE-Core ($W{=}2$, $p{=}0.188$) nor E5+PE-Core ($W{=}1$, $p{=}0.125$) is significant (one-tailed, $\alpha{=}0.05$).

\begin{table}[ht]
\centering

{\setlength{\tabcolsep}{3pt}
\resizebox{\columnwidth}{!}{
\begin{tabular}{lcccc}
\toprule
\textbf{Method} & \textbf{Context} & \textbf{Acc(\%)} & \textbf{F1(\%)} & \textbf{$\Delta_{\text{uni}}$} \\
\midrule
\multicolumn{5}{l}{\textit{Prior published methods}} \\
MULTIBENCH~\cite{PuLiang2021MULTIBENCH:Learning} & \checkmark & 71.8$\pm$0.3 & --- & -- \\
SVM (BERT)~\cite{Castro2019TowardsPaper}     & \checkmark & ---          & 71.5 & -- \\
SWIA~\cite{Chauhan2020SentimentAnalysis}     & \checkmark & ---          & 72.6 & -- \\
CE (BART)~\cite{Ray2022ASarcasm}             & \checkmark & ---          & 74.2 & -- \\
MuLOT (BERT)~\cite{Pramanick2022MultimodalDetection}& \texttimes & 74.52 & ---  & -- \\
IWAN (BERT)~\cite{Wu2021ModelingDetection}   & \texttimes & ---          & 74.5 & -- \\
IWAN (BERT)~\cite{Wu2021ModelingDetection}   & \checkmark & ---          & 75.1 & -- \\
MuLOT (BERT)~\cite{Pramanick2022MultimodalDetection}& \checkmark & 76.82 & ---  & -- \\
MHA (RoBERTa)~\cite{Aggarwal2023MultimodalDataset}     & \checkmark & 79.32        & 77.6 & -- \\
\midrule
\multicolumn{5}{l}{\textit{Our internal baselines -- Audio: Dasheng (base 78.12\%)}} \\
Early Fusion           & \checkmark & 79.71 {\small$\pm$3.04} & 80.0 & \pos{$+1.59$} \\
Late Fusion (Average)  & \checkmark & 79.86 {\small$\pm$4.71} & 80.0 & \pos{$+1.74$} \\
Symmetric Fusion (BERT) & \checkmark & 81.59 {\small$\pm$4.17} & 82.0 & \pos{$+3.47$} \\
Symmetric Fusion (E5)   & \checkmark & 82.46 {\small$\pm$4.52} & 82.0 & \pos{$+4.34$} \\
Proposed (IAF (BERT))  & \checkmark & 82.61 {\small$\pm$3.92} & 83.0 & \pos{$+4.49$} \\
\textbf{Proposed (IAF (E5))}& \checkmark & \best{83.33 {\small$\pm$4.20}} & \best{84.0} & \pos{$+5.21$} \\
\bottomrule
\end{tabular}
}}
\caption{Results on MUStARD. $\Delta_{\text{uni}}$ = gain over Dasheng audio unimodal baseline (our internal baselines only).}
\label{tab:mustard_main}
\end{table}


Table~\ref{tab:pathway_summary} summarises pathway isolation analysis. Across all three datasets, IAF preserves the dominant modality’s pathway at its unimodal baseline, whereas symmetric fusion degrades the same pathways by up to 18.5pp: the E5 text pathway drops from 74.9\% (Table~\ref{tab:unimodal_all}) to 56.4\% under symmetric fusion (Table~\ref{tab:pathway_summary}). This directly supports our claim that IAF’s structural asymmetry functions as a representation-preservation guarantee that symmetric fusion cannot provide. Visual summaries of this collapse pattern are provided in Appendix~\ref{app:collapse_analysis}.

\begin{table}[ht]
\centering

{\setlength{\tabcolsep}{3pt}
\resizebox{\columnwidth}{!}{
\begin{tabular}{l l l c c c c}
\toprule
\textbf{Text} & \textbf{Video} & \textbf{Archi.}
  & \textbf{Full Fusion}
  & \textbf{Text Path}
  & \textbf{Audio Path}
  & \textbf{Video Path} \\
\midrule

\multicolumn{7}{c}{\textit{MultiHuSE -- Dominant Modality: TEXT -- BERT=70.2\%, E5=74.9\%}} \\
\midrule

\multirow{4}{*}{\rotatebox{90}{BERT}}
  & \multirow{2}{*}{VideoMAE}
    & Symmetric & 69.42{\small$\pm$1.48} & 61.3{\small$\pm$2.5} & 59.0{\small$\pm$4.5} & 60.1{\small$\pm$1.3} \\
  & & IAF & 73.16{\small$\pm$2.59} & \best{70.2{\small$\pm$2.8}} & 55.8{\small$\pm$7.4} & 57.5{\small$\pm$5.0} \\

  & \multirow{2}{*}{PE-Core}
    & Symmetric & 72.00{\small$\pm$1.20} & 63.5{\small$\pm$1.6} & 63.6{\small$\pm$2.3} & 61.6{\small$\pm$7.2} \\
  & & IAF & 74.74{\small$\pm$1.71} & \best{70.2{\small$\pm$2.8}} & 62.9{\small$\pm$2.5} & 65.6{\small$\pm$3.1} \\

\midrule

\multirow{4}{*}{\rotatebox{90}{E5}}
  & \multirow{2}{*}{VideoMAE}
    & Symmetric & 72.37{\small$\pm$2.73} & 56.4{\small$\pm$4.6} & 63.7{\small$\pm$9.2} & 67.4{\small$\pm$3.3} \\
  & & IAF & 76.92{\small$\pm$3.53} & \best{74.9{\small$\pm$3.1}} & 48.4{\small$\pm$8.9} & 56.7{\small$\pm$7.0} \\

  & \multirow{2}{*}{PE-Core}
    & Symmetric & 74.32{\small$\pm$3.03} & 55.3{\small$\pm$8.7} & 65.5{\small$\pm$2.2} & 69.7{\small$\pm$4.8} \\
  & & IAF & 78.06{\small$\pm$3.82} & \best{74.9{\small$\pm$3.1}} & 54.9{\small$\pm$9.6} & 62.2{\small$\pm$6.6} \\

\midrule
\multicolumn{7}{c}{\textit{UR-FUNNY -- Dominant Modality: TEXT=65.19\%}} \\
\midrule

\multirow{2}{*}{\rotatebox{90}{GloVe}}
  & \multirow{2}{*}{OpenFace}
    & Symmetric & 68.91 & 63.08 & 49.30 & 50.70 \\
  & & IAF & 70.72 & \best{65.19} & 50.91 & 49.30 \\

\midrule
\multicolumn{7}{c}{\textit{MUStARD-- Dominant Modalities: AUDIO=78.12\%; VIDEO (PE)=77.10\%}} \\
\midrule

\multirow{4}{*}{\rotatebox{90}{BERT}}
  & \multirow{2}{*}{VideoMAE}
    & Symmetric & 79.28{\small$\pm$4.26} & 62.0{\small$\pm$5.0} & 64.4{\small$\pm$15.2} & 67.8{\small$\pm$8.0} \\
  & & IAF & 80.43{\small$\pm$2.63} & 64.6{\small$\pm$9.0} & \best{78.1{\small$\pm$5.5}} & \best{74.5{\small$\pm$1.2}} \\

  & \multirow{2}{*}{PE-Core}
    & Symmetric & 81.59{\small$\pm$4.17} & 64.9{\small$\pm$9.5} & 69.4{\small$\pm$6.3} & 73.5{\small$\pm$9.0} \\
  & & IAF & 82.61{\small$\pm$3.92} & 63.9{\small$\pm$6.2} & \best{78.1{\small$\pm$5.5}} & \best{77.1{\small$\pm$3.1}} \\

\midrule

\multirow{4}{*}{\rotatebox{90}{E5}}
  & \multirow{2}{*}{VideoMAE}
    & Symmetric & 79.86{\small$\pm$3.09} & 60.3{\small$\pm$12.3} & 77.3{\small$\pm$5.1} & 71.5{\small$\pm$6.2} \\
  & & IAF & 79.42{\small$\pm$4.34} & 66.4{\small$\pm$7.6} & \best{78.1{\small$\pm$5.5}} & \best{74.5{\small$\pm$1.2}} \\

  & \multirow{2}{*}{PE-Core}
    & Symmetric & 82.46{\small$\pm$4.52} & 74.4{\small$\pm$3.5} & 74.9{\small$\pm$4.6} & 73.3{\small$\pm$10.4} \\
  & & IAF & 83.33{\small$\pm$4.20} & 69.0{\small$\pm$7.8} & \best{78.1{\small$\pm$5.5}} & \best{77.1{\small$\pm$3.1}} \\

\bottomrule
\end{tabular}
}}
\caption{Pathway isolation analysis. Each score reflects accuracy using only the corresponding modality pathway, bypassing fusion aggregation. \best{Bold} = dominant modality pathway. All metrics in \%.}
\label{tab:pathway_summary}
\end{table}


\subsection{Ablation Study}

\paragraph{(1) Role of Modalities.}
We assess modality contribution through bimodal ablations, reported in Appendix~\ref{app:bimodal_full}. Across all three datasets, full trimodal IAF remains the best configuration. The ablations are consistent with the empirical modality hierarchies identified by unimodal validation: removing text hurts most on MultiHuSE and UR-FUNNY, whereas on MUStARD the text-free A+V setting remains close to full trimodal performance, reflecting the stronger audio-visual signal.

\paragraph{(2) Role of Distilled Feature Representations.}
We contrast IAF trained on raw pre-trained features with IAF trained on distillation-enhanced features, isolating the impact of Stage~1. As seen in Table~\ref{tab:distill_ablation}, incorporating distilled representations improves performance: gains of 1.02–2.41\% on MultiHuSE, 1.02–1.59\% on MUStARD, and 1.81\% on UR-FUNNY. This underscores the importance of raising weaker modality representations before fusion, rather than relying solely on the architecture.

\begin{table}[ht]
\centering

\resizebox{\columnwidth}{!}{
\begin{tabular}{lllccc}
\toprule
\textbf{Dataset} & \textbf{Config} & \textbf{Feat.} &
\textbf{Acc(\%) $\pm$ SD} & \textbf{F1(\%)} & \textbf{$\Delta$}(\%) \\
\midrule

\multirow{4}{*}{MultiHuSE}

  & \multirow{2}{*}{BERT+PEC}
    & Base  & 72.33{\small$\pm$2.02} & 72 & \nega{-2.41} \\
  &      & Dist & \best{74.74{\small$\pm$1.71}} & \best{74} & -- \\
\cmidrule{2-6}

  & \multirow{2}{*}{E5+PEC}
    & Base  & 75.90{\small$\pm$6.17} & 76 & \nega{-2.16} \\
  &      & Dist  & \best{78.06{\small$\pm$3.82}} & \best{78} & -- \\
\midrule

\multirow{4}{*}{MUStARD}

  & \multirow{2}{*}{BERT+PEC}
    & Base  & 81.59{\small$\pm$3.85} & 82 & \nega{-1.02} \\
  &      & Dist  & \best{82.61{\small$\pm$3.92}} & \best{83} & -- \\
\cmidrule{2-6}

  & \multirow{2}{*}{E5+PEC}
    & Base & 81.74{\small$\pm$4.24} & 82 & \nega{-1.59} \\
  &      & Dist  & \best{83.33{\small$\pm$4.20}} & \best{84} & -- \\
\midrule

\multirow{2}{*}{UR-FUNNY}
  & \multirow{2}{*}{GloVe/COV/OF}
    & Base  & 68.91 & 69 & \nega{-1.81} \\
  &      & Dist & \best{70.72} & \best{71} & -- \\
\bottomrule
\end{tabular}
}
\caption{Distillation ablation: raw (Base) vs.\ distilled (Dist.) features within IAF. 
$\Delta$ = Dist $-$ Base.}
\label{tab:distill_ablation}
\end{table}

\paragraph{(3) Role of Regularisation.}
Appendix~\ref{app:reg_ablation}, Table~\ref{tab:combined_reg_ablation}, reports the contribution of Mixup and Curriculum Modality Dropout. Removing both yields the largest drops across all datasets (up to 4.07\% on MultiHuSE, 1.61\% on UR-FUNNY, and 2.90\% on MUStARD), confirming their complementary roles. Curriculum Dropout provides a consistently larger isolated gain than Mixup, suggesting that progressive difficulty scheduling is particularly effective in noisy multimodal feature spaces.

\paragraph{(4) Role of the Gate Mechanism.}
We evaluate three gating strategies within IAF: a \textit{Learnable} gate
(sample-adaptive weights), a \textit{Hierarchy} gate (fixed weights derived from modality ranking), and a \textit{Uniform} gate (equal weights, T=A=V=0.33). Results (Appendix~\ref{ablation:gates}, 
Table~\ref{tab:gate_ablation}) shows that the learnable gate generally performs best, while the hierarchy gate remains competitive under strong modality imbalance. Appendix~\ref{app:gate_weights} further shows that mean gate weights remain distributed across modalities in all configurations, with no modality exceeding 0.53 on average.

\section{Conclusion}
We introduced Inverted Asymmetric Fusion (IAF), a two-stage multimodal framework for mitigating strong-modality collapse in discriminative fusion. By preserving the dominant modality through a \textsc{pure}/\textsc{attend} hierarchy and strengthening weaker modalities via knowledge distillation prior to fusion, IAF maintains the dominant pathway at its unimodal ceiling while improving fused performance in text-dominant and audio-visual-dominant settings. Across three benchmarks with different modality hierarchies, IAF consistently outperforms early, late, and symmetric fusion baselines, and surpasses prior published methods on UR-FUNNY and MUStARD. These findings suggest that protecting strong modalities from cross-modal interference is an important architectural principle for robust multimodal learning.

\section*{Limitations}
This work has several limitations. First, IAF requires the dominant modality to be identified empirically through unimodal evaluation before fusion training. When modalities have similar predictive strength, the \textsc{pure}/\textsc{attend} assignment becomes less clear and the benefits of structural asymmetry may diminish. Moreover, the inferred hierarchy depends on the encoder set: a stronger text encoder may appear dominant over a weaker video encoder even when the task itself relies more heavily on visual information. The hierarchy therefore reflects empirical encoder performance rather than intrinsic task structure. Although the dual-teacher design used for MUStARD partially addresses co-dominant settings, a more principled treatment remains open, for example through dynamic hierarchy assignment at the sample level.

Second, all experiments use frozen, pre-extracted features. This isolates the architectural contribution of IAF from encoder fine-tuning, but the framework has not been evaluated end to end, where joint optimisation could alter both the modality hierarchy and the effectiveness of distillation. In addition, our pathway analysis uses the original frozen unimodal classifier as a fixed probe, so it tests whether the dominant modality's decision boundary is preserved rather than whether it remains recoverable in principle. Future work could examine this more directly through probing or representational analyses such as Centred Kernel Alignment (CKA).

Finally, all datasets are English-only and confined to humor and sarcasm. It therefore remains unclear whether the modality hierarchies observed here, and the benefits of IAF, generalise to other languages or to tasks such as emotion recognition, visual question answering, or action recognition. Although the inclusion of E5-multilingual suggests potential cross-lingual applicability, this has not been evaluated, and cross-dataset generalisation remains for future work.

\section*{Ethical considerations}
This work uses three publicly available datasets (MultiHuSE, UR-FUNNY, MUStARD), collected and released under standard academic licences by their respective authors. No new data collection or human annotation was conducted. All experiments operate on pre-extracted, anonymised feature representations with no access to personally identifiable information. We note that automated humor and sarcasm detection systems may misclassify culturally specific communication styles and are not intended for deployment beyond controlled research settings.

\section*{Acknowledgments}
This research was supported by the Petroleum Technology Development Fund (PTDF) of Nigeria.

\bibliography{references}


\appendix
\vspace{-1em}
\section{Full Hyperparameter Configurations}
\label{app:hyperparams}
\enlargethispage{3\baselineskip}
Table~\ref{tab:hyperparams_distill} and Table~\ref{tab:hyperparams_fusion} present the complete hyperparameter configurations for the distillation and fusion stages, across all datasets. Hyperparameters were selected through manual tuning: values were adjusted iteratively based on validation performance until stable, well-performing configurations were identified, after which a single final experiment was run per configuration and reported in the paper. No automated search was conducted.

\begin{table}[h]
\centering

\setlength{\tabcolsep}{3pt} 
    \renewcommand{\arraystretch}{0.5} 
    \vspace{-0.5em}
\resizebox{\columnwidth}{!}{
\begin{tabular}{lccc}
\toprule
\textbf{Hyperparameter} & \textbf{MultiHuSE} & \textbf{UR-FUNNY} & \textbf{MUStARD} \\
\midrule
Optimiser               & AdamW              & Adam              & AdamW            \\
Learning rate           & $6 \times 10^{-5}$ & $1 \times 10^{-5}$ & $1 \times 10^{-3}$ \\
Weight decay            & $10^{-3}$          & $10^{-3}$         & $10^{-2}$        \\
Max epochs              & 100                & 80                & 100              \\
Early stopping patience & 35                 & 20                & 40               \\
Gradient clipping       & 1.0                & ---               & 1.0              \\
\bottomrule
\end{tabular}
}

\caption{Dataset-specific hyperparameters for the fusion}
\label{tab:hyperparams_fusion}
\end{table}

\begin{table}[ht]
    \centering
    {\small 
    \setlength{\tabcolsep}{3pt} 
    \renewcommand{\arraystretch}{0.5} 
    \vspace{-0.5em}
    \begin{tabular}{lc}
    \toprule
    \textbf{Hyperparameter} & \textbf{Value} \\
    \midrule
    KD loss weights $(\alpha,\beta,\gamma)$ & 0.4, 0.35, 0.25 \\
    Temperature $T$ & 3.5 \\
    Optimiser & AdamW \\
    Weight decay & $10^{-3}$ \\
    Text/Audio LR & $10^{-4}$--$10^{-3}$ \\
    Video LR & $9\times10^{-5}$ \\
    \bottomrule
\end{tabular}
\caption{Distillation stage hyperparameters.}
\label{tab:hyperparams_distill}
}
\end{table} 

\section{Role of Modalities Ablation}
\label{app:bimodal_full}

We remove one modality at a time to quantify each modality's contribution to full trimodal performance. Table~\ref{tab:bimodal_ablation} shows that full trimodal IAF outperforms all bimodal variants on every dataset. On MultiHuSE, the text-free A+V pair drops to 65.85\%, confirming text’s dominant role. On UR-FUNNY, $T{+}A$ outperforms $T{+}V$ (68.91\% vs.\ 65.29\%), consistent with the Text $>$ Audio $\approx$ Video hierarchy. On MUStARD, the text-free A+V pair (82.61\%) nearly matches full trimodal performance (83.33\%), reflecting the co-dominance of audio and video and the smaller marginal benefit of text.

\begin{table}[ht]
\centering

\resizebox{\columnwidth}{!}{
\begin{tabular}{l c c c c}
\toprule
\textbf{Configuration} 
  & \textbf{Pure} & \textbf{Attends} 
  & \textbf{Acc(\%) $\pm$ SD} & \textbf{F1(\%)} \\
\midrule

\multicolumn{5}{c}{\textit{MultiHuSE — Text dominant (E5, Dasheng, PE-Core)}} \\
\midrule
Full Trimodal          & Text  & Audio, Video & \best{78.06 {\small$\pm$3.82}} & \best{78} \\
T+V                & Text  & Video        & 76.61 {\small$\pm$4.58} & 77 \\
T+A                & Text  & Audio        & 76.57 {\small$\pm$2.92} & 76 \\
A+V  (Text-Free)   & Audio & Video        & 65.85 {\small$\pm$1.63} & 65 \\

\midrule
\multicolumn{5}{c}{\textit{UR-FUNNY — Text dominant (GloVe, COVAREP, OpenFace)}} \\
\midrule
Full Trimodal          & Text  & Audio, Video & \best{70.72} & \best{71} \\
T+A                & Text  & Audio        & 68.91        & 69 \\
T+V                & Text  & Video        & 65.29        & 65 \\
A+V (Text-Free)   & Audio & Video        & 65.29         & 65 \\

\midrule
\multicolumn{5}{c}{\textit{MUStARD — Audio/Video dominant (E5, Dasheng, PE-Core)}} \\
\midrule
Full Trimodal          & Audio, Video & Text & \best{83.33 {\small$\pm$4.20}} & \best{84} \\
A+T                & Audio        & Text & 79.57 {\small$\pm$5.17} & 80 \\
V+T                & Video        & Text & 79.57 {\small$\pm$5.41} & 80 \\
A+V  (Text-Free)   & Audio        & Video& 82.61 {\small$\pm$3.01} & 83 \\

\bottomrule
\end{tabular}
}
\caption{Bimodal ablation of E5 and GloVe-based configuration for IAF  }
\label{tab:bimodal_ablation}
\end{table}

Table~\ref{tab:bimodal_bert} reports bimodal ablation results for BERT encoder configurations on MultiHuSE and MUStARD. Results follow the same directional patterns as the E5 configurations in Table~\ref{tab:bimodal_ablation}, confirming that the modality hierarchy findings are consistent across text encoders.

\begin{table}[ht]
\centering

{\setlength{\tabcolsep}{3pt}
\resizebox{\columnwidth}{!}{
\begin{tabular}{l c c c c}
\toprule
\textbf{Configuration} & \textbf{Pure} & \textbf{Attends} & \textbf{Acc (\%) $\pm$ SD} & \textbf{F1}(\%) \\
\midrule

\multicolumn{5}{c}{\textit{MultiHuSE — Text dominant (BERT, Dasheng, PE-Core)}} \\
\midrule
Full Trimodal          & Text & Audio, Video & \best{74.74 {\small{}$\pm$ 1.71}} & \best{74} \\
T+V                & Text & Video        & 72.45 {\small{}$\pm$ 3.27} & 72 \\
T+A                & Text & Audio        & 72.08 {\small{}$\pm$ 2.71} & 72 \\

\midrule
\multicolumn{5}{c}{\textit{MUStARD — Audio/Video dominant (BERT, Dasheng, PE-Core)}} \\
\midrule
Full Trimodal          & Audio, Video & Text & \best{82.61 {\small{}$\pm$ 3.92}} & \best{83} \\
A+T                & Audio        & Text & 79.71 {\small{}$\pm$ 3.30} & 80 \\
V+T                & Video        & Text & 77.25 {\small{}$\pm$ 4.75} & 77 \\
\bottomrule
\end{tabular}
}}
\caption{Bimodal fusion for BERT configurations on MultiHuSE and MUStARD.}
\label{tab:bimodal_bert}
\end{table}

\section{Role of Regularisation Ablation}
\label{app:reg_ablation}

Table~\ref{tab:combined_reg_ablation} reports the full regularisation ablation across all datasets and encoder configurations. BERT- and E5-based configurations are shown for MultiHuSE and MUStARD; results are directionally consistent across both encoders.

\begin{table}[ht]
\centering

{\small
\setlength{\tabcolsep}{4pt} 
    \renewcommand{\arraystretch}{0.5} 
\resizebox{\columnwidth}{!}{
\begin{tabular}{ccccc}
\toprule
\textbf{Mixup} & \textbf{Curriculum} & \textbf{Acc (\%) $\pm$ SD} & \textbf{F1} & \textbf{$\Delta$ (\%)} \\

\midrule
\multicolumn{5}{c}{\textbf{MultiHuSE} \textit{(BERT + Dasheng + PE-Core)}} \\
\midrule
\ding{51} & \ding{51} & \best{74.74 {\small$\pm$1.71}} & \best{74} & \textbf{--} \\
\ding{55} & \ding{51} & 73.49 {\small$\pm$3.15} & 73 & \nega{$-1.25$} \\
\ding{51} & \ding{55} & 72.58 {\small$\pm$3.80} & 72 & \nega{$-2.16$} \\
\ding{55} & \ding{55} & 70.75 {\small$\pm$2.67} & 70 & \nega{$-3.99$} \\

\midrule
\multicolumn{5}{c}{\textbf{MultiHuSE} \textit{(E5 + Dasheng + PE-Core)}} \\
\midrule
\ding{51} & \ding{51} & \best{78.06 {\small$\pm$3.82}} & \best{78} & \textbf{--} \\
\ding{55} & \ding{51} & 76.61 {\small$\pm$4.40} & 77 & \nega{$-1.45$} \\
\ding{51} & \ding{55} & 75.74 {\small$\pm$5.53} & 76 & \nega{$-2.32$} \\
\ding{55} & \ding{55} & 73.99 {\small$\pm$5.04} & 74 & \nega{$-4.07$} \\

\midrule
\multicolumn{5}{c}{\textbf{UR-FUNNY} \textit{(GloVe + COVAREP + OpenFace)}} \\
\midrule
\ding{51} & \ding{51} & \best{70.72} & \best{71} & \textbf{--} \\
\ding{51} & \ding{55} & 70.62 & 71 & \nega{$-0.10$} \\
\ding{55} & \ding{51} & 69.72 & 70 & \nega{$-1.00$} \\
\ding{55} & \ding{55} & 69.11 & 69 & \nega{$-1.61$} \\

\midrule
\multicolumn{5}{c}{\textbf{MUStARD} \textit{(BERT + Dasheng + PE-Core)}} \\
\midrule
\ding{51} & \ding{51} & \best{82.61 {\small$\pm$3.92}} & \best{83} & \textbf{--} \\
\ding{51} & \ding{55} & 82.03 {\small$\pm$4.08} & 82 & \nega{$-0.58$} \\
\ding{55} & \ding{51} & 81.45 {\small$\pm$4.11} & 81 & \nega{$-1.16$} \\
\ding{55} & \ding{55} & 79.71 {\small$\pm$4.25} & 80 & \nega{$-2.90$} \\

\midrule
\multicolumn{5}{c}{\textbf{MUStARD} \textit{(E5 + Dasheng + PE-Core)}} \\
\midrule
\ding{51} & \ding{51} & \best{83.33 {\small$\pm$4.20}} & \best{84} & \textbf{--} \\
\ding{51} & \ding{55} & 82.03 {\small$\pm$3.73} & 82 & \nega{$-1.30$} \\
\ding{55} & \ding{51} & 81.59 {\small$\pm$3.65} & 82 & \nega{$-1.74$} \\
\ding{55} & \ding{55} & 81.16 {\small$\pm$4.51} & 81 & \nega{$-2.17$} \\
\bottomrule
\end{tabular}
}
\caption{Regularisation ablation across MultiHuSE, UR-FUNNY, and MUStARD. \ding{51}\ding{51} = both Mixup and Curriculum Dropout enabled (full model); $\Delta$ = accuracy change relative to the full model. All metrics in \%.}
\label{tab:combined_reg_ablation}
}
\end{table}

\section{Role of Gate Mechanism Ablation}
\label{ablation:gates}
Table~\ref{tab:gate_ablation} reports the effect of different gating strategies. The learnable gate attains the highest accuracy on UR-FUNNY and MUStARD, demonstrating the benefit of sample-adaptive weighting. MultiHuSE is a mild exception where the fixed hierarchy gate performs marginally better, likely due to its extreme modality imbalance: under highly skewed conditions, a learnable gate can over-emphasize the dominant modality. Overall differences between gates are small ($\leq$1.45\%), indicating that IAF is largely insensitive to the precise gating strategy.

\begin{table}[t]
\centering

\setlength{\tabcolsep}{4pt} 
\renewcommand{\arraystretch}{0.9} 

\resizebox{\columnwidth}{!}{
\begin{tabular}{l l c c}
\toprule
\textbf{Config} & \textbf{Gate Type} & \textbf{Acc(\%) $\pm$ SD} & \textbf{$\Delta$}(\%) \\
\midrule

\multicolumn{4}{c}{\textit{MultiHuSE Dataset} (Hierarchy: T=0.50, A=0.30, V=0.20)} \\
\midrule

\multirow{3}{*}{BERT + PE-Core}
  & Learnable  & 74.74 {\small$\pm$1.71} & \textbf{--} \\
  & Hierarchy  & \best{74.99 {\small$\pm$2.68}} & \pos{$+0.25$} \\
  & Uniform    & 74.86 {\small$\pm$3.31} & \pos{$+0.12$} \\
\cmidrule{1-4}

\multirow{3}{*}{E5 + PE-Core}
  & Learnable  & 78.06 {\small$\pm$3.82} & \textbf{--} \\
  & Hierarchy  & \best{78.52 {\small$\pm$3.64}} & \pos{$+0.46$} \\
  & Uniform    & 78.02 {\small$\pm$3.81} & \nega{$-0.04$} \\

\midrule
\multicolumn{4}{c}{\textit{UR-FUNNY Dataset} (Hierarchy: T=0.50, A=0.30, V=0.20)} \\
\midrule

\multirow{3}{*}{GloVe/COVAREP/OpenFace}
  & Learnable  & \best{70.72} & \textbf{--} \\
  & Hierarchy  & 70.32 & \nega{$-0.40$} \\
  & Uniform    & 69.42 & \nega{$-1.30$} \\

\midrule
\multicolumn{4}{c}{\textit{MUStARD Dataset} (Hierarchy: T=0.25, A=0.40, V=0.35)} \\
\midrule

\multirow{3}{*}{BERT + PE-Core}
  & Learnable  & \best{82.61 {\small$\pm$3.92}} & \textbf{--} \\
  & Hierarchy  & 81.74 {\small$\pm$4.11} & \nega{$-0.87$} \\
  & Uniform    & 81.88 {\small$\pm$3.94} & \nega{$-0.73$} \\
\cmidrule{1-4}

\multirow{3}{*}{E5 + PE-Core}
  & Learnable  & \best{83.33 {\small$\pm$4.20}} & \textbf{--} \\
  & Hierarchy  & 82.03 {\small$\pm$4.48} & \nega{$-1.30$} \\
  & Uniform    & 81.88 {\small$\pm$4.42} & \nega{$-1.45$} \\

\bottomrule
\end{tabular}
}
\caption{Gate mechanism ablation across all three datasets. Learnable = sample-adaptive weights; Hierarchy = fixed weights derived from modality ranking; Uniform = equal weights (T=A=V=0.33). $\Delta$ = accuracy difference relative to the Learnable gate.}
\label{tab:gate_ablation}
\end{table}

\section{Pathway Collapse: Visualisation}
\label{app:collapse_analysis}

This section visualises pathway collapse in the distilled IAF and symmetric models (E5 + PE-Core, MultiHuSE) from two complementary perspectives: aggregate pathway accuracy across configurations and feature-space behaviour of the text pathway.

\begin{figure}[ht]
  \centering
  \includegraphics[width=\columnwidth]{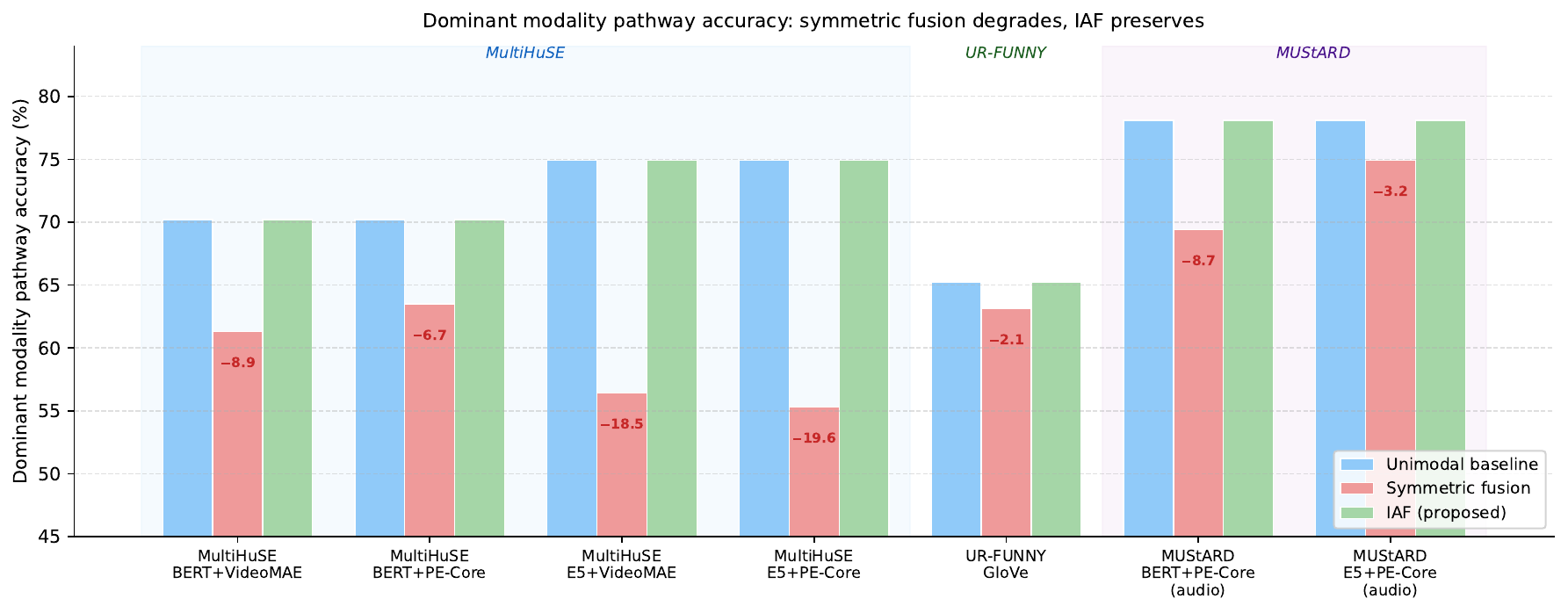}
  \caption{Dominant-modality pathway accuracy for all configurations (data from Table~\ref{tab:pathway_summary}). Shaded regions group MultiHuSE (blue) and MUStARD (purple) configurations.}
  \label{fig:collapse_barchart}
\end{figure}

Figure~\ref{fig:collapse_barchart} summarises dominant-modality pathway accuracy for all configurations in Table~\ref{tab:pathway_summary}, comparing the unimodal ceiling, symmetric fusion, and IAF. Across all seven settings, symmetric fusion consistently reduces pathway accuracy, with a mean drop of 11.6\,pp, whereas IAF matches the unimodal ceiling in every case. This pattern holds across datasets and encoder combinations, suggesting that strong-modality collapse is a structural property of symmetric cross-attention fusion rather than an artefact of any particular encoder.
 
\begin{figure}[ht]
  \centering
  \includegraphics[width=\columnwidth]{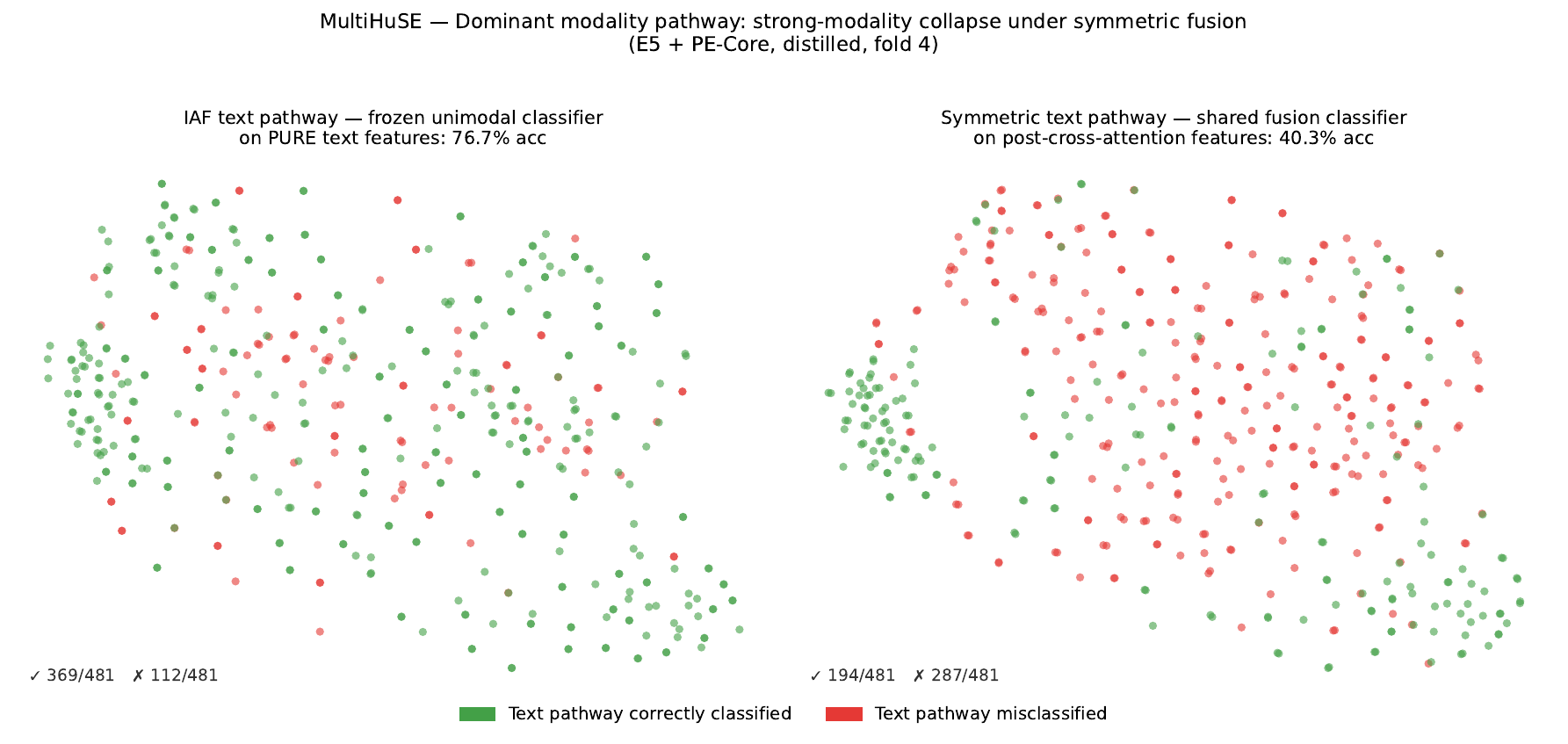}
  \caption{Joint t-SNE of features used at the final classification step for the text pathway under IAF (\textit{left}) and symmetric fusion (\textit{right}), fold~4 of MultiHuSE (E5\,+\,PE-Core, distilled). \textcolor[HTML]{43A047}{Green} = correctly classified by that pathway; \textcolor[HTML]{E53935}{red} = misclassified.}
  \label{fig:tsne_collapse}
\end{figure}

\paragraph{Feature-space visualisation of collapse (t-SNE).}
Figure~\ref{fig:tsne_collapse} shows a joint t-SNE embedding of the text-pathway features used at the final classification step, coloured by prediction correctness. Both panels share a single embedding fitted on the combined feature matrix, so differences in error density reflect classifier behaviour rather than changes in the underlying feature geometry.

In the IAF panel (left), the frozen unimodal classifier $\psi_T$ operates on the backbone features $\bm{h}^{(T)}$ on which it was trained; text-pathway accuracy on fold~4 is 76.7\%. In the symmetric fusion panel (right), the shared classifier operates on post-cross-attention features $\tilde{\bm{h}}^{(T)}$; accuracy drops to 40.3\%. The residual connection preserves a broadly similar manifold structure across both panels, yet the symmetric pathway produces substantially more errors in the same geometric space. This pattern is consistent with classifier mismatch: the shared head is not optimised for single-pathway inputs.
 
\section{Gate Weight Analysis}
\label{app:gate_weights}

The IAF gate network produces per-sample weights $(w_T, w_A, w_V)$ that sum to one and determine each modality’s contribution before classification. Table~\ref{tab:gate_weights_all} reports mean gate weights averaged across cross-validation folds at the best checkpoint for all nine configurations. Two population-level patterns emerge. First, the modality assigned the \textsc{pure} role consistently receives the highest mean weight: text leads on MultiHuSE ($\bar{w}_T \in [0.35, 0.47]$), whereas audio and video dominate on MUStARD ($\bar{w}_A + \bar{w}_V > 0.80$ in all configurations). On UR-FUNNY, weights are more evenly distributed ($\bar{w}_T = 0.372$, $\bar{w}_A = 0.310$, $\bar{w}_V = 0.318$), consistent with more balanced cross-modal cues in that corpus. Second, no modality exceeds a mean weight of 0.53 in any configuration, suggesting that IAF preserves multimodal integration rather than collapsing into a regularised unimodal model.

\begin{table}[ht]
\centering
\setlength{\tabcolsep}{6pt}
\renewcommand{\arraystretch}{1.1}
\resizebox{\columnwidth}{!}{%
\begin{tabular}{llcccc}
\toprule
\textbf{Text} & \textbf{Feat.} & \makecell{$\boldsymbol{\Delta}$\textbf{(\%)}} & $\bar{w}_T$ & $\bar{w}_A$ & $\bar{w}_V$ \\
\midrule
\multicolumn{6}{c}{\textbf{MultiHuSE}
  \textit{(Text \textsc{pure} $|$ Dasheng+PE-Core)}} \\
\midrule
\multirow{2}{*}{BERT+PE-Core}
  & Base  & $+4.07^{[\text{B}]}$ & 0.474 & 0.355 & 0.171 \\
  & Dist. & $+6.48^{[\text{B}]}$ & 0.355 & 0.206 & 0.439 \\
\multirow{2}{*}{E5+PE-Core}
  & Base  & $+1.04^{[\text{E}]}$ & 0.431 & 0.425 & 0.144 \\
  & Dist. & $+3.20^{[\text{E}]}$ & 0.412 & 0.197 & 0.391 \\
\midrule
\multicolumn{6}{c}{\textbf{UR-FUNNY}
  \textit{(Text \textsc{pure} $|$ COVAREP + OpenFace)}} \\
\midrule
GloVe & Dist. & $+8.25$ & 0.372 & 0.310 & 0.318 \\
\midrule
\multicolumn{6}{c}{\textbf{MUStARD}
  \textit{(Audio+Video \textsc{pure} $|$ Text \textsc{attend})}} \\
\midrule
\multirow{2}{*}{BERT+PE-Core}
  & Base  & $+3.47^{[\text{B}]}$ & 0.194 & 0.530 & 0.276 \\
  & Dist. & $+4.49^{[\text{B}]}$ & 0.161 & 0.389 & 0.449 \\
\multirow{2}{*}{E5+PE-Core}
  & Base  & $+3.62^{[\text{E}]}$ & 0.166 & 0.440 & 0.394 \\
  & Dist. & $+5.21^{[\text{E}]}$ & 0.103 & 0.470 & 0.428 \\
\bottomrule
\end{tabular}
}
\caption{Mean learned gate weights across CV folds at the best checkpoint. $\Delta$ = gain over dominant unimodal baseline (MultiHuSE: BERT\,=\,68.26\%, E5\,=\,74.86\%; UR-FUNNY: GloVe\,=\,62.47\%; MUStARD: Dasheng\,=\,78.12\%). \textsuperscript{[B]}/\textsuperscript{[E]}: gap from BERT/E5 baseline. Base = raw pre-trained; Dist.\ = knowledge-distilled.}
\label{tab:gate_weights_all}
\end{table}

These averages, however, mask substantial within-dataset variation. Table~\ref{tab:gate_case_study} presents representative samples from the distilled IAF model (E5\,+\,PE-Core) on MultiHuSE. Although the population average is $\bar{w}_T \approx 0.51$, individual samples range from near-exclusive text weighting ($w_T > 0.95$) to strong audio dominance ($w_A > 0.85$). Four cases illustrate this adaptive behaviour. In Category~(i), unambiguous headline-style utterances labelled as neutral humour are classified primarily from text. In Category~(ii), deadpan or context-dependent phrasing provides weak textual evidence, and the gate shifts weight toward audio and visual cues while preserving correct classification. In Category~(iii), the model shows over-reliance on text: utterances such as \textit{Age is of no importance unless you're a cheese} depend on delivery cues absent from the transcript, leading to misclassification. In Category~(iv), audio becomes dominant when the transcript is relatively uninformative; for example, \textit{Tomorrow is 2-22-22. Happy Tuesday!} appears textually plain, but vocal warmth and intonation support affiliative classification, and the gate assigns about 90\% of the weight to audio. Overall, the aggregate trends align with the \textsc{pure}/\textsc{attend} assignments, while the case studies show that the gate adapts to sample-level variation in modality reliability.

\begin{table}[ht]
\centering
\small
\setlength{\tabcolsep}{4pt}
\renewcommand{\arraystretch}{1.15}
\resizebox{\columnwidth}{!}{
\begin{tabular}{p{5.2cm}cccccl}
\toprule
\textbf{Sample text} & \textbf{True} & \textbf{Pred} & $w_T$ & $w_A$ & $w_V$ & \\
\midrule
\multicolumn{7}{l}{\textit{(i) High text weight, correctly classified}} \\[2pt]
Joe Kennedy III reveals how his GOP\newline counterparts really feel about Trump's tweets. & Neu & Neu & 0.970 & 0.006 & 0.024 & \ding{51} \\
Katy Perry wears American flag dress to\newline kids in overall concerts. & Neu & Neu & 0.961 & 0.018 & 0.020 & \ding{51} \\
Porn actress confirms Trump affair after\newline unpublished 2011 interview. & Neu & Neu & 0.959 & 0.020 & 0.021 & \ding{51} \\
\midrule
\multicolumn{7}{l}{\textit{(ii) Low text weight, correctly classified}} \\[2pt]
The only positive thing about you is your HIV status. & Agg & Agg & 0.023 & 0.548 & 0.429 & \ding{51} \\
Advice to ice skaters: you can't always\newline tell a brick by its cover. & Neu & Neu & 0.024 & 0.643 & 0.333 & \ding{51} \\
He who laughs last probably doesn't\newline understand the joke. & Neu & Neu & 0.028 & 0.685 & 0.287 & \ding{51} \\
\midrule
\multicolumn{7}{l}{\textit{(iii) High text weight, incorrectly classified}} \\[2pt]
Age is of no importance unless you're a cheese. & S-enh & Agg & 0.885 & 0.032 & 0.082 & \ding{55} \\
Don't look back. You're not going that way. & S-enh & Agg & 0.863 & 0.049 & 0.088 & \ding{55} \\
What is the best contraceptive for old people? Nudity. & Agg & Aff & 0.857 & 0.084 & 0.058 & \ding{55} \\
\midrule
\multicolumn{7}{l}{\textit{(iv) Audio-dominant, correctly classified}} \\[2pt]
How many Californians does it take to screw in a light bulb? None. They screw in hot tubs. & Neu & Neu & 0.039 & 0.900 & 0.061 & \ding{51} \\
Tomorrow is 2-22-22. Happy Tuesday! & Aff & Aff & 0.135 & 0.856 & 0.009 & \ding{51} \\
Why couldn't the leopard play hide and seek? Because he was always spotted. & Aff & Aff & 0.102 & 0.851 & 0.048 & \ding{51} \\
\bottomrule
\end{tabular}
}
\caption{Representative IAF gate weight profiles (E5\,+\,PE-Core, MultiHuSE, all folds).\textcolor{red}{\textit{ This table contains potentially offensive or sensitive example text reproduced from the dataset for analysis.} }Class labels: Aff\,=\,Affiliative, Agg\,=\,Aggressive, Neu\,=\,Neutral, S-enh\,=\,Self-enhancing. $w_T{+}w_A{+}w_V=1$ per sample. \ding{51}~correct; \ding{55}~incorrect.}
\label{tab:gate_case_study}
\end{table}

\section{Distillation Coefficient Sensitivity}
\label{app:sensitivity}

Table~\ref{tab:hyp_sensitivity} reports the effect of varying the three distillation loss coefficients (\(\alpha\): hard CE, \(\beta\): soft KD, \(\gamma\): feature alignment) on MultiHuSE. We evaluate three alternative configurations alongside the default setting: equal weights, a CE-heavy variant, and a soft-target-heavy variant. All distilled students are then used as inputs to IAF under the same fusion training procedure.

Fusion accuracy for IAF with E5, audio, and PE-Core varies by at most 1.20 percentage points across the four configurations. The two best-performing settings, the default and equal-weight variants, achieve the same top-line accuracy of 78.06\%. CE-heavy weighting yields a small improvement in video distillation (+0.38\,pp relative to the default) but reduces cross-modal soft transfer, leading to a 1.20\,pp drop in fusion accuracy.

\begin{table}[ht]
\centering
\small
\setlength{\tabcolsep}{3.5pt}
\resizebox{\columnwidth}{!}{%
\begin{tabular}{lcccccc}
\toprule
\textbf{Setting} & \textbf{Coeff.\ ($\alpha$/$\beta$/$\gamma$)}
  & \textbf{BERT Dist.} & \textbf{Audio Dist.} & \textbf{Video Dist.}
  & \textbf{IAF (BERT)} & \textbf{IAF (E5)} \\
\midrule
Default$^\dagger$
  & 0.40 / 0.35 / 0.25
  & \textbf{70.21}\small{$\pm$2.82} & 62.90\small{$\pm$1.83} & 45.57\small{$\pm$1.22}
  & \textbf{74.74}\small{$\pm$1.71} & \textbf{78.06}\small{$\pm$3.82} \\
Equal weights
  & 0.33 / 0.33 / 0.33
  & 69.92\small{$\pm$2.79} & \textbf{63.11}\small{$\pm$1.54} & 44.66\small{$\pm$1.57}
  & 74.37\small{$\pm$2.33} & \textbf{78.06}\small{$\pm$3.36} \\
CE-heavy
  & 0.60 / 0.25 / 0.15
  & 69.63\small{$\pm$2.75} & 62.78\small{$\pm$1.18} & \textbf{45.95}\small{$\pm$0.98}
  & 74.03\small{$\pm$2.48} & 76.86\small{$\pm$4.32} \\
Soft-target-heavy
  & 0.25 / 0.55 / 0.20
  & 70.17\small{$\pm$2.58} & 62.86\small{$\pm$1.10} & 44.74\small{$\pm$1.35}
  & 73.33\small{$\pm$2.32} & 76.90\small{$\pm$4.18} \\
\midrule
Range & -- & 0.58 & 0.33 & 1.29 & 1.41 & 1.20 \\
\bottomrule
\end{tabular}
}
\caption{Distillation coefficient sensitivity on MultiHuSE (5-fold CV, mean\,$\pm$\,SD\,\%).  $\dagger$\,=\,setting used in all main experiments.  IAF columns use distilled students; Audio\,=\,Dasheng, Video\,=\,PE-Core throughout. ``Range'' reports max\,$-$\,min across the four settings.}
\label{tab:hyp_sensitivity}
\end{table}

The soft-target-heavy variant lowers video distillation performance (-0.83\,pp relative to the default), suggesting that the weakest modality depends more on direct task supervision through CE than on soft targets from a teacher trained on a different modality. In this setting, increasing the weight on soft targets from the text teacher appears to overwhelm the video student's task-specific learning signal. Audio distillation remains stable across all settings, with a range of at most 0.33\,pp, indicating that the Dasheng encoder adapts consistently regardless of coefficient choice. Overall, these results suggest that IAF is not highly sensitive to the precise coefficient values; the default setting was selected by manual tuning and then held fixed across all datasets.

\section{Knowledge Distillation Design Comparison}
\label{app:kd_comparison}

Table~\ref{tab:kd_comparison} situates MAKD relative to prior multimodal distillation approaches along five design axes: training stage, teacher source, objective, and optimisation procedure. The key distinction is that MAKD is a standalone pre-fusion stage that uses only the empirically dominant unimodal modality as teacher to strengthen weaker encoders before fusion, whereas prior multimodal KD methods typically operate within the fusion stage itself.

\begin{table}[t]
\centering
\setlength{\tabcolsep}{3pt}
\resizebox{\columnwidth}{!}{%
\begin{tabular}{lllll}
\toprule
\textbf{Method} & \textbf{Stage} & \textbf{Teacher} & \textbf{Purpose} & \textbf{Procedure} \\
\midrule
\makecell[l]{Cross-M KD \cite{Albanie2018EmotionWild} \\ \cite{Thoker2019Cross-ModalRecognition}}
  & Pre-fusion$^{\dagger}$ & Paired modality   & Unimodal transfer          & Sequential \\
Wang et al.~\cite{Wang2020MultimodalDistillation}
  & Within fusion          & Full multimodal   & Dropout robustness         & Joint      \\
MMANet~\cite{Wei2023MMANet:Learning}
  & Within fusion          & Full multimodal   & Missing-modality & Joint      \\
DMD~\cite{Li2023DecoupledRecognition}
  & Within fusion          & Subspace teachers & Repr.\ disentanglement     & Concurrent \\
MTMD~\cite{Lin2024Multi-TaskAnalysis}
  & Within fusion          & Unimodal + fused  & Heterogeneity reduction    & Concurrent \\
\midrule
\textbf{MAKD (Ours)}
  & \textbf{Pre-fusion}    & \textbf{Dom.\ unimodal} & \textbf{Encoder strengthening} & \textbf{Sequential} \\
\bottomrule
\end{tabular}%
}
\caption{Design comparison of multimodal knowledge distillation methods.
$^{\dagger}$Not designed in the context of a subsequent fusion stage.}
\label{tab:kd_comparison}
\end{table}

\end{document}

%% file: methodology_figure.tex
\begin{tikzpicture}[
  >=Stealth, font=\small,
  encbox/.style={draw=#1, fill=#1!10, rounded corners=5pt,
    minimum width=1.6cm, minimum height=0.58cm,
    text=#1!80!black, font=\footnotesize\bfseries, align=center},
  featvec/.style={draw=#1!70, fill=#1!25, rounded corners=2pt,
    minimum width=0.30cm, minimum height=1.0cm},
  teacherrect/.style={draw=KDcol, fill=KDcol!8, rounded corners=6pt,
    minimum width=2.0cm, minimum height=0.65cm,
    font=\footnotesize\bfseries, text=KDcol!80!black, align=center},
  studentrect/.style={draw=#1!80, fill=#1!8, rounded corners=5pt,
    minimum width=1.55cm, minimum height=0.58cm,
    font=\footnotesize, text=#1!80!black, align=center},
  lossbox/.style={draw=KDcol!50, fill=KDcol!5, rounded corners=5pt,
    minimum width=2.2cm, minimum height=1.05cm,
    font=\scriptsize, align=left},
  purepath/.style={draw=PUREbdr, fill=PUREfill, rounded corners=5pt,
    minimum width=2.2cm, minimum height=0.62cm,
    font=\footnotesize\bfseries, text=PUREbdr!80!black, align=center},
  cmapath/.style={draw=ATTbdr, fill=ATTfill, rounded corners=5pt,
    minimum width=2.2cm, minimum height=0.62cm,
    font=\footnotesize\bfseries, text=ATTbdr!80!black, align=center},
  psibox/.style={draw=#1!65, fill=#1!8, rounded corners=4pt,
    minimum width=1.1cm, minimum height=0.55cm,
    font=\scriptsize, align=center},
  gatebox/.style={draw=GATEcol!55, fill=GATEcol!5, rounded corners=5pt,
    minimum width=4.6cm, minimum height=0.68cm,
    font=\scriptsize, align=center, text=GATEcol},
  wsumbox/.style={draw=GATEcol!45, fill=GATEcol!4, rounded corners=5pt,
    minimum width=3.2cm, minimum height=0.65cm,
    font=\footnotesize, align=center},
  outbox/.style={draw=gray!55, fill=white, rounded corners=6pt,
    minimum width=1.6cm, minimum height=0.72cm,
    font=\footnotesize\bfseries, align=center,
    blur shadow={shadow blur steps=5,shadow xshift=1pt,
                 shadow yshift=-1pt,shadow opacity=20}},
  sectiontitle/.style={fill=#1, text=white, rounded corners=4pt,
    font=\footnotesize\bfseries, minimum width=2.8cm,
    minimum height=0.46cm, align=center},
  panelbg/.style={rounded corners=9pt, fill=panelbg,
    draw=sepline, inner sep=9pt},
  mainarrow/.style={->, thick, #1},
  puresignal/.style={->, line width=2.0pt, Tcol},
  distillarrow/.style={->, thick, KDcol, dashed},
  ctxarrow/.style={->, thin, Tcol!50, dashed},
  gatearrow/.style={->, thin, GATEcol!40, dotted},
]

\node[sectiontitle=Tcol!75!black] (s1t) at (1.2,4.4) {Unimodal Encoders};

\node[encbox=Tcol] (tenc) at (0.3,2.8) {E5 / BERT};
\node[featvec=Tcol, right=0.3cm of tenc] (tfv) {};
\node[font=\tiny\bfseries,text=Tcol,below=2pt of tfv]{$\mathbf{h}^{(t)}$};
\draw[mainarrow=Tcol](tenc)--(tfv);
\node[font=\tiny\itshape,text=Tcol!65,right=2pt of tfv,yshift=5pt]{dominant};

\node[encbox=Acol] (aenc) at (0.3,1.2) {Dasheng};
\node[featvec=Acol, right=0.3cm of aenc] (afv) {};
\node[font=\tiny\bfseries,text=Acol,below=2pt of afv]{$\mathbf{h}^{(a)}$};
\draw[mainarrow=Acol](aenc)--(afv);

\node[encbox=Vcol] (venc) at (0.3,-0.4) {PE-Core};
\node[featvec=Vcol, right=0.3cm of venc] (vfv) {};
\node[font=\tiny\bfseries,text=Vcol,below=2pt of vfv]{$\mathbf{h}^{(v)}$};
\draw[mainarrow=Vcol](venc)--(vfv);

\node[draw=gray!25,fill=gray!5,rounded corners=3pt,
      font=\tiny,align=center,text=gray!55,minimum width=2.0cm]
      (acclabel) at (1.2,-1.60)
      {$\mathrm{Acc}(m^*)>\mathrm{Acc}(m{\neq}m^*)$};

\begin{scope}[on background layer]
\node[panelbg,fit=(s1t)(tenc)(venc)(vfv)(acclabel),inner sep=10pt]{};
\end{scope}

\node[sectiontitle=KDcol!80] (s2t) at (5.0,4.4) {Stage 1:\\[-1pt]Knowledge Distillation};

\node[teacherrect] (teacher) at (5.0,3.0) {Teacher $(m^*)$};

\node[lossbox] (losseq) at (5.0,1.75)
      {$\mathcal{L}_{\mathrm{KD}}=\alpha\mathcal{L}_{\mathrm{CE}}$\\[1pt]
       $\quad{+}\,\beta\mathcal{L}_{\mathrm{soft}}$\\[1pt]
       $\quad{+}\,\gamma\mathcal{L}_{\mathrm{feat}}$};

\node[studentrect=Acol] (astud) at (3.9,0.1) {Audio\\[-1pt]Student};
\node[studentrect=Vcol] (vstud) at (6.1,0.1) {Video\\[-1pt]Student};

\draw[distillarrow](teacher)--(losseq)
      node[midway,right=2pt,font=\tiny,text=KDcol]{distil};
\draw[distillarrow](losseq.south west)--(astud.north);
\draw[distillarrow](losseq.south east)--(vstud.north);

\node[font=\small\itshape,text=KDcol!55,align=center]
      (kdnote) at (5.0,-1.6)
      {Direction adapts to\\dataset modality hierarchy};

\draw[mainarrow=Tcol,dashed]
      (tfv.east)--++(0.2,0)..controls+(1.2,0)and+(-1.5,0)..(teacher.west);
\draw[mainarrow=Acol,dashed]
      (afv.east)--++(0.2,0)..controls+(0.8,0)and+(-1.2,0)..(astud.west);
\draw[mainarrow=Vcol,dashed]
      (vfv.east)--++(0.2,0)..controls+(3.8,0.5)and+(-0.4,-1)..(vstud.south);

\begin{scope}[on background layer]
\node[panelbg,fit=(s2t)(teacher)(losseq)(astud)(vstud)(kdnote),inner sep=10pt]{};
\end{scope}

\node[sectiontitle=ATTbdr!85!black] (s3t) at (10.0,4.4)
      {Stage 2:\\[-1pt]Inverted Asymmetric Fusion};

\node[purepath] (pure) at (9.8,3.0)
      {\textsc{pure}\;$\mathbf{o}^{(m^*)}=\mathbf{h}^{(m^*)}$};
\draw[puresignal]
      (teacher.east)--++(0.2,0)..controls+(1.5,0)and+(-2.0,0)..(pure.west);

\node[cmapath] (cma_a) at (9.8,1.2) {\textsc{attend}\;|\;CMA};
\node[cmapath] (cma_v) at (9.8,-0.6) {\textsc{attend}\;|\;CMA};

\draw[mainarrow=Acol]
      (astud.east)--++(0.2,0)..controls+(0.8,0)and+(-4.8,0)..(cma_a.west);
\draw[mainarrow=Vcol]
      (vstud.east)--++(0.2,0)..controls+(0.8,0)and+(-1.8,0)..(cma_v.west);

\draw[ctxarrow](pure.south)--++(0,-0.28)-|(cma_a.north)
      node[near end,right=2pt,font=\tiny,text=Tcol!50]{context};
\draw[ctxarrow](pure.south)--++(0,-0.5)-|(cma_v.north);

\node[psibox=Tcol] (psi_t) at (12.4,3.0) {$\psi_{m^*}$\\\tiny frozen};
\node[psibox=Acol] (psi_a) at (12.4,1.2) {$\psi_a$\\\tiny trainable};
\node[psibox=Vcol] (psi_v) at (12.4,-0.6) {$\psi_v$\\\tiny trainable};

\draw[mainarrow=Tcol](pure.east)--(psi_t.west);
\draw[mainarrow=Acol](cma_a.east)--(psi_a.west);
\draw[mainarrow=Vcol](cma_v.east)--(psi_v.west);

\node[gatebox] (gate) at (10.3,-1.6)
      {Gate\;$\mathcal{G}$:\;
       $\mathbf{w}=\mathrm{softmax}\!\left(
         \mathcal{G}([\mathbf{h}^{(t)}\|\mathbf{h}^{(a)}\|\mathbf{h}^{(v)}])
       \right)\in\mathbb{R}^3$};

\draw[gatearrow](tfv.south)--++(0,-0.35)
      ..controls+(5.5,-1.8)and+(-4.5,0.6)..(gate.west);
\draw[gatearrow](afv.south)--++(0,-0.25)
      ..controls+(5.5,-1.0)and+(-4.5,0.2)..(gate.west);
\draw[gatearrow](vfv.south)--++(0,-0.15)
      ..controls+(5.5,-0.4)and+(-4.5,-0.2)..(gate.west);

\node[draw=regcol!35,fill=regcol!5,rounded corners=5pt,
      font=\tiny,align=center,text=regcol,
      minimum width=4.6cm,minimum height=0.62cm]
      (reg) at (10.3,-2.4)
      {\textbf{Training Regularisation:}\;
       Mixup $(\alpha{=}0.05)$\;+\;Curriculum Modality Dropout};

\begin{scope}[on background layer]
\node[panelbg,fit=(s3t)(pure)(cma_v)(psi_t)(psi_v)(gate)(reg),inner sep=10pt]{};
\end{scope}

\node[wsumbox] (wsum) at (15.5,1.2)
      {$\hat{\mathbf{y}} = \sum_{m} w_m\cdot\psi_m(\mathbf{o}^{(m)})$};

\draw[->,thick](psi_t.east)--++(0.15,0)-|(wsum.north);
\draw[->,thick](psi_a.east)--(wsum.west);
\draw[->,thick](psi_v.east)--++(0.15,0)-|(wsum.south);
\draw[->,GATEcol!45,thick,dashed](gate.east)--++(0.25,0)-|(wsum.south);

\node[outbox] (out) at (18.5,1.2) {Prediction\\[1pt]$\hat{\mathbf{y}}$};
\draw[->,line width=1.6pt](wsum.east)--(out.west);

\node[font=\tiny,text=gray!55,align=center,below=5pt of out]
      {\textit{Humor style} (MultiHuSE)\\
       \textit{Humor\,/\,Sarcasm}\\(UR-FUNNY, MUStARD)};

\draw[sepline,dashed,thin](2.9,-2.75)--(2.9,4.6);
\draw[sepline,dashed,thin](7.6,-2.75)--(7.6,4.6);
\draw[sepline,dashed,thin](13.5,-2.75)--(13.5,4.6);

\node[font=\tiny\bfseries,text=gray!65] at (0.6,-3.2) {Modalities:};
\foreach \lbl/\col/\xpos in {
  Text/Tcol/2.0, Audio/Acol/3.2, Video/Vcol/4.4}{
  \node[draw=\col,fill=\col!15,rounded corners=2pt,font=\tiny,
        minimum width=0.9cm,minimum height=0.27cm] at (\xpos,-3.2) {\lbl};
}
\node[draw=PUREbdr,fill=PUREfill,rounded corners=2pt,font=\tiny,
      minimum width=1.55cm,minimum height=0.27cm] at (6.3,-3.2)
      {\textsc{pure}: no cross-attn};
\node[draw=ATTbdr,fill=ATTfill,rounded corners=2pt,font=\tiny,
      minimum width=1.65cm,minimum height=0.27cm] at (8.8,-3.2)
      {\textsc{attend}: queries all};
\node[draw=KDcol!55,fill=KDcol!8,rounded corners=2pt,font=\tiny,
      minimum width=1.4cm,minimum height=0.27cm] at (11.2,-3.2)
      {Knowledge Distil.};
\draw[puresignal](12.3,-3.2)--++(0.8,0);
\node[font=\tiny,text=Tcol] at (14.3,-3.2) {Dominant (preserved)};

\end{tikzpicture}